\documentclass{article}

\usepackage{arxiv}

\usepackage[T1]{fontenc}
\usepackage[utf8]{inputenc}

\usepackage{hyperref}
\usepackage{url}
\usepackage{booktabs}
\usepackage{amsmath}      
\usepackage{amsfonts}
\usepackage{amssymb}
\usepackage{nicefrac}
\usepackage{microtype}
\usepackage{cleveref}
\usepackage{graphicx}
\usepackage[round]{natbib}
\usepackage{doi}

\usepackage{subcaption}
\usepackage{multirow}
\usepackage{xcolor}

\newcommand{\E}{\mathbf{E}}
\newcommand{\R}{\mathbb{R}}

\newcommand{\charDim}{d_c}
\newcommand{\posDim}{d_p}

\newcommand{\dmodel}{d_{\text{model}}}

\title{Kronecker Embeddings:\\Byte-Level Structured Token Representations\\for Parameter-Efficient Language Models}

\author{
  Rohan Shravan \\
  The School of AI \\
  Bengaluru, India \\
  \texttt{rshravan@theschoolofai.in}
}

\renewcommand{\shorttitle}{Kronecker Embeddings}

\hypersetup{
  hidelinks,                        
  pdftitle={Kronecker Embeddings: Byte-Level Structured Token Representations for Parameter-Efficient Language Models},
  pdfauthor={Rohan Shravan},
  pdfkeywords={language models, embeddings, byte-level, parameter efficiency, tokenization},
}

\begin{document}
\maketitle

\begin{abstract}
Large language models route every input through a learned embedding table
of shape $|V| \times \dmodel$. At frontier scale this consumes hundreds of
millions to billions of trainable parameters before any contextual
computation. We introduce Kronecker Embeddings, a deterministic byte-level
character-position factorization that replaces this table with a fixed
encoder and a single learned projection, preserving compatibility with
standard BPE tokenizers while eliminating 91--94\% of input-side trainable
parameters at frontier scale.

We provide five contributions. First, a cross-model probe across
six modern LMs (135M to 671B parameters) shows that trained input embeddings
cluster typographic variants of the probe word (\verb|run| $\to$
\verb|Run|, \verb| run|, \verb|.run|) far more than morphological relatives,
across tokenizer families and roughly five orders of magnitude of
training compute. Kronecker embeddings escape this typographic
clustering at the embedding layer. A layered probe of our own trained 124M checkpoints
further shows that at this scale neither method's nearest-neighbor geometry
develops strict morphological clustering in the first two transformer
layers; the BPE arm's first layer trades morphological geometry for
co-occurrence/contextual geometry while Kronecker's geometry is preserved
through early layers. Whether these layered findings hold at frontier scale
is an open question for future work.

Second, a controlled training comparison on the standard nanoGPT GPT-2 124M
architecture trained on 2.5B tokens of FineWeb-Edu shows Kronecker
reaching $2.5 \pm 0.2\%$ lower validation loss than the BPE-tied baseline
($n=3$ seeds, gap $0.083 \pm 0.007$ nats, approximately 9\% lower
validation perplexity), with the gap widening through training and
stabilizing at convergence rather than narrowing. Kronecker requires
approximately $1.43\times$ fewer optimizer steps to reach BPE's converged
loss.

Third, a behavioral spelling-robustness probe on the same trained
checkpoints across 110 (clean, typo) prompt pairs shows that Kronecker's
predictions are more robust to single-character typographical errors on
every aggregate metric we measured: top-1 predicted token preserved on
55.5\% of pairs vs.\ 47.3\% for BPE ($+$8.2\,pp), mean
KL$(\text{clean}\|\text{typo})$ 0.79 vs.\ 0.86 ($-$7.6\%),
final hidden-state cosine 0.879 vs.\ 0.857 ($+$2.6\%). Kronecker wins or ties
top-1 stability in 10 of 11 prompt categories. A complementary
qualitative generation probe finds that the Kronecker arm
\emph{echoes} byte-novel strings and misspellings through autoregressive
generation (preserving \verb|kronekticus| and \verb|netwrok| through
30-token continuations) where the BPE arm fragments and forgets them.

Fourth, a measured mechanism observation: BPE embedding norm walks from
std$=0.020$ to $0.026$ across training while Kronecker projection norm
stays at $\sim 1.0$ throughout, consistent with Kronecker providing a
stable representational target for the transformer body.

Fifth, we describe an on-the-fly runtime variant that reconstructs
embeddings from a 4.5\,MB byte buffer rather than a 2.15\,GB
precomputed table at production vocabulary size 131{,}072, with the
recomputation overhead measured at 0.01--0.24\% of step time on
internal 9B and 120B-MoE configurations.

The method has tradeoffs: byte-level locality means semantically distant
but byte-similar pairs cluster together (\verb|compute|/\verb|commute|,
\verb|nation|/\verb|notion|), shifting some disambiguation work to the
transformer's first attention layers.

\end{abstract}


\vspace{0.5em}
\noindent\textbf{Code and reproducibility.} A reference implementation of
Kronecker Embeddings (codec, precomputed and on-the-fly variants,
\texttt{nn.Embedding}-compatible interface), the probe scripts used to
produce the empirical results in Section~\ref{sec:results}, and the
raw output JSON files for the spelling-robustness and generation probes
are released at \url{https://github.com/theschoolofai/kronecker-embeddings}
under the Apache~2.0 license.

\vspace{0.25em}
\noindent\textbf{Released models.} Four LightningLM 0.1V models
(2B dense, 5B-MoE, 9B-MoE, and 120B-MoE), which use Kronecker
embeddings as their input layer, are released at
\url{https://huggingface.co/theschoolofai}. These are complete
architectures incorporating several novel components beyond the
embedding layer and are described in separate work; they are not
controlled embedding ablations. The controlled isolation of the
embedding contribution is the three-seed 124M comparison of
Section~\ref{sec:results-nanogpt}. We release the LightningLM models
as evidence that Kronecker embeddings support stable, convergent
training in a full modern architecture at scale, in the same sense
that a mechanism validated in a controlled small-scale study is
subsequently deployed in a larger system.

\section{Introduction}
\label{sec:intro}

\subsection{The token embedding bottleneck}
\label{sec:intro-bottleneck}

The input pipeline of a modern transformer language model begins with a
single matrix lookup. Each token identifier from a vocabulary $V$ is mapped
to a continuous vector through a learned embedding table
$\E \in \R^{|V| \times \dmodel}$. The table is a parameter, optimized
jointly with the rest of the network during pretraining.

At small scale, this matrix is cheap. A 135M-parameter model with a 50K
vocabulary and $\dmodel = 768$ allocates roughly $38\text{M}$ parameters
to its input embedding, a quarter of the total parameter count, but still
modest in absolute terms. The arithmetic changes at frontier scale. A
130K-vocabulary model with $\dmodel = 4096$ allocates $537\text{M}$
parameters to its input embedding alone. A hypothetical 250K-vocabulary
multilingual model with $\dmodel = 24576$ allocates $6.14\text{B}$
parameters to its input embedding, an order of magnitude larger than the
entire transformer body of an 8B model. These parameters carry the usual
optimizer-state overhead: in mixed-precision Adam \citep{kingma2015adam}
training with bf16 weights and fp32 master copies plus first and
second moments, each trainable parameter costs approximately 16 bytes
of optimizer state.

Beyond the raw count, the embedding table introduces engineering frictions
specific to large training runs. The table must be sharded across data-parallel
ranks. Gradients must be all-reduced or sharded. In tensor-parallel setups
the table is split along the vocabulary axis, requiring vocabulary-wise
communication on every forward and backward pass. Loading and saving
checkpoints involves transferring a multi-gigabyte tensor that is not
participating in any non-trivial computation. It is a fixed mapping from
token identities to vectors, and we ask training to slowly learn what that
fixed mapping should be.

This paper asks whether the learned mapping is necessary.

\subsection{What structure do input embeddings actually develop?}
\label{sec:intro-structure}

A common informal view holds that trained input embeddings develop
semantic and morphological structure: that
$\E[\text{``run''}]$ and $\E[\text{``running''}]$ should cluster in the
sense that morphologically related words occupy nearby points in
embedding space, and that the transformer leverages this geometry to
generalize.

We tested this view across six modern public language models
(Llama-3.2-1B, Qwen3-32B, Gemma-3-1B-pt, DeepSeek-V3-Base, GPT-OSS-120B,
SmolLM2-135M; collectively spanning 135M to 671B parameters and three
tokenizer families). Our findings are summarized here and developed in
detail in Section~\ref{sec:results}.

\textbf{Finding 1: trained embeddings cluster typographic variants.}
For a probe word $s$, the nearest neighbors of $\E[s]$ in centered cosine
distance are dominated by typographic variants of $s$ itself: the same
word with different case, with leading whitespace, or with leading
punctuation. The pattern is consistent across models, tokenizer families,
and roughly five orders of magnitude of training compute. Even
DeepSeek-V3-Base, trained on 14.8T tokens (the most-trained open model in
our set), retrieves \verb|run|, \verb| run|, \verb|Run|, \verb| Run|,
\verb|.run| as its top-5 neighbors of \verb|run|. The morphological
relatives \verb|running|, \verb|runner|, \verb|ran| are not in the top-K.

\textbf{Finding 2: Kronecker embeddings escape typographic clustering.}
A deterministic byte-level encoder that maps each token to a Kronecker
product of byte-and-position basis vectors produces embeddings whose
nearest neighbors are byte-similar strings: \verb|run|'s neighbors include
\verb|runs|, \verb|rund|, \verb|runner|, \verb|ru|, \verb|ron|. Some of
these are morphological family members; others are byte-similar
non-relatives. The escape from typographic clustering is robust at the
embedding layer.

\textbf{Finding 3: neither method robustly captures strict morphology.}
When morphological-relatedness is measured strictly (against a hand-curated
list of family members rather than the looser "different canonical form"
criterion), both methods retrieve fewer than 30\% genuine morphological
family members in their top-10 neighborhoods at the embedding layer. We
also find that the first two transformer layers of a controlled 124M
training comparison do not build strict morphological clustering on top of
either embedding: in the BPE arm, embedding-level morphological structure
\emph{declines} through layers as the transformer trades it for
co-occurrence and context; in the Kronecker arm, byte-level geometry is
preserved largely unchanged. We caution that this layered finding is at
small scale (124M GPT-2 trained on 2.5B tokens) and may not generalize to
frontier-scale models, where larger transformers may construct
early-layer morphological structure that our small-model probe cannot
detect. Replicating the layered probe at 1B+ scale is a clear follow-up.

Taken together, these findings suggest that the two embedding schemes
provide \emph{different inductive biases at the input layer}: trained BPE
clusters by co-occurrence-shaped typographic identity, while Kronecker
clusters by byte-level surface similarity. The controlled training
comparison in Section~\ref{sec:results-nanogpt} measures which prior
yields better validation loss when both are paired with the same
transformer body.

\subsection{What this paper proposes}
\label{sec:intro-proposal}

We propose Kronecker Embeddings: a structured replacement for the learned
input embedding table. Each token's vector is computed as the Kronecker
product of one-hot byte representations and one-hot byte-position
representations, summed across the bytes of the token's UTF-8 surface form,
length-normalized, and projected to $\dmodel$ through a single learned
linear map. The byte-and-position encoder is fixed; the projection is
trainable.

The pipeline is a drop-in replacement for \verb|nn.Embedding|. It accepts
the same input (token identifiers from a standard BPE
\citep{sennrich2016neural} or SentencePiece \citep{kudo2018sentencepiece}
tokenizer) and produces the same output (a $\dmodel$-dimensional vector
per token). All other architectural choices (attention, MLP, layer norm,
positional encoding, loss) remain unchanged. The replacement reduces the
input-side trainable parameter count by 91--94\% across the scales we
characterize (Table~\ref{tab:param-savings}). We emphasize that these
figures refer to the input side only. Because the Kronecker codec
dimension $D \neq \dmodel$ in general, weight tying is architecturally
inapplicable (Section~\ref{sec:method-output-head}), so a tied-BPE
baseline at small scale loses its shared input/output matrix when
replaced by Kronecker: the output head reappears as a separate
trainable block of the same size as the original tied embedding.
The net trainable-parameter comparison at 124M scale is given
explicitly in Section~\ref{sec:results-runtime}.

\paragraph{Reallocated parameters, not eliminated parameters.}
A useful reframing: the parameters Kronecker removes from the input
embedding are not lost, they are reallocated. At fixed total parameter
budget, the transformer body absorbs the savings: more attention heads,
wider MLPs, more layers, or larger MoE expert pools. The 35M parameters
that would have been spent at the 124M scale memorizing a
token-id-to-vector mapping become available for representational work
in the body. At the 120B-MoE scale the same accounting frees
$\sim$503M parameters ($\sim$94\% of the input side) for the body.
Equivalently, at fixed body size, the total trainable parameter count
drops by that amount, which reduces optimizer-state memory pressure on
the remaining parameters (Adam-style optimizers carry roughly 8 bytes
of state per trainable parameter on top of the parameter itself;
eliminating 503M trainable parameters frees roughly 4\,GB of optimizer
state per data-parallel rank in addition to the weight memory savings).

\paragraph{No embedding table to ship.} A second consequence is that
deployed Kronecker-trained models do not need to carry their input
embedding table as a shipped artifact. The deployed artifact for a
Kronecker-trained model consists of the transformer body weights, the
learned $D \to \dmodel$ projection matrix, and the tokenizer
configuration (which already encodes the token-id-to-bytes mapping
needed to compute the codec on the fly). The byte buffer itself is
either re-derived from the tokenizer at load time or shipped as a
few-megabyte file. This is particularly consequential for edge
deployment and for low-precision-quantized models; see
Section~\ref{sec:results-shipping} for concrete numbers.

\subsection{Five concrete contributions}
\label{sec:intro-contributions}

\begin{enumerate}
\item \textbf{The method.} Definition, mathematical formulation,
implementation. The Kronecker codec, the per-token length normalization,
the learned projection, and the on-the-fly runtime variant.

\item \textbf{Cross-model probe.} Empirical characterization of what
structure trained input embeddings actually develop across six modern LMs
spanning four labs and three tokenizer families. We find typographic
clustering, not morphological clustering, at the embedding layer; we
introduce a metric (\emph{loose morph@K}) that quantifies escape from
typographic clustering.

\item \textbf{Controlled training comparison.} A three-seed comparison on
the standard nanoGPT GPT-2 124M architecture trained on 2.5B tokens of
FineWeb-Edu, in which only the input embedding scheme differs between
arms. Kronecker reaches $2.5 \pm 0.2\%$ lower validation loss than
BPE-tied baseline, with the gap stable across seeds and through
convergence. A companion 138M custom-architecture run reproduces the
direction of result.

\item \textbf{Spelling-robustness and byte-level fidelity.} On a 110-pair
clean/typo robustness probe, Kronecker preserves the top-1 prediction
on 55.5\% of pairs vs.\ 47.3\% for BPE ($+$8.2 percentage points), wins
or ties top-1 stability in 10 of 11 categories, and reduces KL
divergence between clean and typo distributions by 7.6\%. A qualitative
generation probe shows that the Kronecker arm echoes byte-novel strings
and misspellings through autoregressive generation while BPE fragments
them.

\item \textbf{Mechanism observation and on-the-fly runtime.} We measure a
representational stability effect: BPE embedding norm drifts during
training while Kronecker projection norm remains stable, supporting the
hypothesis of a fixed-target representational regime. We additionally
describe an on-the-fly Kronecker variant that exchanges fixed compute for
embedding-table memory; at production scale (vocab 131,072, $D=8192$) the
runtime variant stores a 4.5\,MB byte buffer instead of a 2.15\,GB
precomputed table, with measured overhead of 0.01--0.24\% of step time on
internal 9B and 120B-MoE configurations.
\end{enumerate}

\subsection{What this paper is not}
\label{sec:intro-not}

This paper is not a tokenizer paper; we use standard BPE or SentencePiece
tokenization, unchanged. It is not a character-level model; the
transformer processes BPE token sequences, not bytes or characters. It is
not a tokenization-free approach in the style of CANINE
\citep{clark2022canine} or ByT5 \citep{xue2022byt5}; we do not eliminate
the tokenizer or operate on byte sequences for the transformer's input.
The work targets exactly one component (the input embedding
table) and proposes a structured replacement that retains tokenization
and modifies nothing else.


\section{Background and Related Work}
\label{sec:background}

\subsection{Learned token embeddings}
\label{sec:bg-learned}

Modern transformer language models begin with a token-id-to-vector
lookup: an embedding matrix $\E \in \R^{|V| \times \dmodel}$ is
parameter-initialized (typically Gaussian or scaled-uniform with small
standard deviation) and learned jointly with the rest of the network
\citep{vaswani2017attention}. The matrix grows linearly in vocabulary
size and model width, and at frontier scale dominates input-side
parameter accounting. Section~\ref{sec:intro-bottleneck} works through
the arithmetic.

\subsection{Weight tying}
\label{sec:bg-tying}

\citet{press2017using} introduced the practice of tying the input
embedding matrix to the output projection matrix: $\E^{\top} = W_{\text{out}}$.
The argument was empirical: tying improved perplexity in their
RNN-LM experiments and reduced the parameter count. Tying remains
standard in many small-to-medium LMs (Llama-3.2, Gemma-3, SmolLM2).

\citet{lopardo2026weighttying} recently showed that tied embeddings are
biased toward the output prediction space rather than the input
representation space: the shared matrix aligns more closely with the
\emph{output} (unembedding) matrices of comparable untied models than with
the \emph{input} embeddings, indicating that tying systematically pulls
the embedding toward output-prediction geometry. The mechanism, they
argue, is gradient asymmetry during training: the output gradient
dominates the shared matrix early.

Several frontier models have moved away from tying: DeepSeek-V3-Base,
Qwen3-32B, and OLMo-2 all use untied embeddings at scale. Our cross-model
probe (Section~\ref{sec:results-stability}) includes both tied and
untied models among the six trained LMs and finds the typographic
clustering pattern in both, suggesting the inductive bias is not specific
to tying.

Because Kronecker's projection dimension $D$ generally differs from
$\dmodel$, weight tying is architecturally inapplicable to the Kronecker
input pathway. We discuss this in
Section~\ref{sec:method-output-head}.

\subsection{Factorized embeddings}
\label{sec:bg-albert}

\citet{lan2019albert} proposed factorizing the embedding matrix in ALBERT
as $\E = V E_F$, where $V \in \R^{|V| \times E}$ and
$E_F \in \R^{E \times \dmodel}$ with $E \ll \dmodel$. This reduces the
parameter count from $|V| \cdot \dmodel$ to $|V| \cdot E + E \cdot \dmodel$.
Both factor matrices are learned.

Kronecker Embeddings can be viewed as a factorization in the same
spirit, with two differences: the first factor (the byte-and-position
encoder) is deterministic rather than learned, and the factorization is
structured by the byte composition of the token's surface form rather
than by an arbitrary latent dimension. The result is a smaller learned
matrix and an inductive bias at initialization
(Section~\ref{sec:pretraining}) that ALBERT-style factorization
does not provide.

\subsection{Character-aware models}
\label{sec:bg-charcnn}

\citet{kim2016character} introduced character-aware language models
combining a character-level CNN over the bytes of each token, a highway
network, and an LSTM body. The architecture demonstrated that
character-level representations of token surface forms could match or
exceed word-level baselines while using a fraction of the parameters,
particularly on morphologically rich languages.

The relationship to our work is conceptual: both approaches replace a
learned word/token embedding table with a deterministic-or-structured
function of the token's surface form. The differences are
implementation-level: \citet{kim2016character} use a CNN+highway, we use
a structured Kronecker factorization; they targeted RNN-era LMs with
word-level tokenization, we target transformers with BPE/SentencePiece
tokenization.

\subsection{Tokenization-free byte and character models}
\label{sec:bg-tokfree}

The broader history of open-vocabulary modeling and the trade-offs
between word, subword, character, and byte-level granularities is
surveyed by \citet{mielke2021between}; we summarize only the strands
most directly relevant here.

CANINE \citep{clark2022canine} and ByT5 \citep{xue2022byt5} eliminate the
tokenizer entirely and operate on raw Unicode codepoints or UTF-8 bytes.
Charformer \citep{tay2022charformer} learns latent subword segmentation
end-to-end via a gradient-based block-scoring module operating on
bytes; MEGABYTE \citep{yu2023megabyte} introduces a multi-scale
architecture predicting million-byte sequences with patch-level
attention; the Byte Latent Transformer
\citep{pagnoni2024blt} dynamically segments bytes into entropy-based
patches and matches tokenization-based LLMs at scale. MYTE
\citep{limisiewicz2024myte} constructs a morphology-driven byte
encoding aimed specifically at multilingual fairness.
These models pay a substantial sequence-length cost (a 200-token
English sentence becomes a 1000-byte sequence) which they offset through
downsampling, segmentation, or architectural changes.

Our work is \emph{not} tokenization-free. We retain a standard BPE or
SentencePiece tokenizer and operate on token sequences of the same
length the tokenizer produces; we only replace the embedding lookup
inside the model. This preserves the sequence-length efficiency of
subword tokenization while gaining a byte-level inductive bias at the
input layer. ByT5's empirical finding that byte-level models are more
robust to noise and spelling variation
\citep{xue2022byt5} is consistent with our argument for byte-level
locality (Section~\ref{sec:method-locality}) but the implementation
strategy differs.

\subsection{Embedding-space anisotropy}
\label{sec:bg-anisotropy}

\citet{mu2018allbutthetop} observed that trained word embeddings exhibit
substantial anisotropy: there is a dominant mean direction and a few
top principal components shared across all word vectors, encoding
frequency rather than semantic content. They proposed
\emph{all-but-the-top} postprocessing (subtracting the mean and removing
top principal components) and showed it improves downstream task
performance. \citet{gao2019representation} described a similar
phenomenon, ``representation degeneration,'' in trained natural language
generation models and proposed regularization to mitigate it.
\citet{ethayarajh2019contextual} extended the analysis to
transformer-era contextualized embeddings (BERT, ELMo, GPT-2),
showing anisotropy persists across layers and that upper layers
produce more context-specific representations than lower layers.

Our anisotropy diagnostic in Section~\ref{sec:results-stability}
extends these observations to six modern frontier-scale LMs. The
finding that GPT-OSS-120B has $\sim$$60\times$ more anisotropy than
the next most-anisotropic model in our set is, to our knowledge, novel.
Throughout our probes we mean-center before computing cosine
similarities, following \citet{mu2018allbutthetop}.

\subsection{Where Kronecker Embeddings sit}
\label{sec:bg-positioning}

Kronecker Embeddings occupy a specific niche in the design space:
\begin{itemize}
\item \textbf{Versus learned embeddings:} replace a large learned matrix
with a fixed structured encoder and a small learned projection.
\item \textbf{Versus tied embeddings:} compatible with untied output
heads only; the architecture sidesteps the tying-related biases
identified by \citet{lopardo2026weighttying}.
\item \textbf{Versus ALBERT-style factorization \citep{lan2019albert}:}
deterministic rather than learned first factor; built around byte-level
composition of token surface forms.
\item \textbf{Versus CharCNN \citep{kim2016character}:} structured
Kronecker factorization rather than CNN; integrated into a standard
BPE-tokenized transformer rather than a character-level LSTM.
\item \textbf{Versus CANINE/ByT5
\citep{clark2022canine,xue2022byt5}:} preserves tokenization and
sequence-length efficiency; only the embedding lookup is replaced.
\end{itemize}

The narrow scope (``replace the input embedding table, change nothing
else'') is intentional. It allows the controlled comparisons of
Section~\ref{sec:results} to isolate the effect of the embedding scheme
from confounds in tokenization, sequence handling, or architectural
modifications.


\section{Method}
\label{sec:method}

\subsection{System overview}
\label{sec:method-overview}

Kronecker Embeddings replaces the input embedding pathway of a standard
transformer language model. The replacement accepts a sequence of token
identifiers (from any BPE-style or SentencePiece tokenizer) and produces
a sequence of $\dmodel$-dimensional vectors, the same input that the
transformer body would otherwise consume.

The pathway consists of three steps:
\begin{enumerate}
\item For each token id $i$, look up the token's UTF-8 byte sequence
$b_i = (b_{i,1}, \ldots, b_{i,L_i})$ from the tokenizer (which already
knows this mapping).
\item Compute a fixed Kronecker codec vector
$\kappa_i \in \R^{D}$, where $D = \charDim \cdot \posDim$,
encoding the byte-and-position composition of $b_i$.
\item Apply a single learned linear projection
$\mathbf{W}_{\text{proj}} \in \R^{D \times \dmodel}$:
$\mathbf{e}_i = \kappa_i \cdot \mathbf{W}_{\text{proj}}$.
\end{enumerate}

The codec is fixed (no gradient). The projection is the only trainable
parameter on the input side. The output $\mathbf{e}_i$ has the same
shape and feeds the same downstream components as a standard learned
embedding.

\subsection{The codec}
\label{sec:method-codec}

The Kronecker codec is the centerpiece. For a token with byte sequence
$b = (b_1, \ldots, b_L)$ of length $L \leq \posDim$:

\begin{equation}
\label{eq:codec}
\kappa(b) = \frac{1}{\sqrt{L}} \sum_{p=1}^{L}
    \mathbf{c}_{b_p} \otimes \mathbf{p}_p
\end{equation}

where $\mathbf{c}_{b_p} \in \R^{\charDim}$ is a one-hot vector encoding
the byte value at position $p$ (so $\mathbf{c}_v$ has a 1 in
coordinate $v$ and zeros elsewhere; we use $\charDim = 256$, the full
byte alphabet), and $\mathbf{p}_p \in \R^{\posDim}$ is a one-hot vector
encoding the byte position $p$ within the token
($\mathbf{p}_p$ has a 1 in coordinate $p$). The Kronecker product
$\mathbf{c}_{b_p} \otimes \mathbf{p}_p \in \R^{\charDim \cdot \posDim}$
is the byte-position basis vector for $(b_p, p)$; it has a single 1 in
coordinate $b_p \cdot \posDim + p$ and zeros elsewhere.

The codec $\kappa$ is a deterministic function of the byte sequence. Its
output is a $D$-dimensional vector with at most $L$ nonzero coordinates,
each equal to $1/\sqrt{L}$.

\paragraph{Length normalization.} The $1/\sqrt{L}$ factor is a
variance-preserving normalization: under any reasonable byte distribution, the
expected squared L2 norm of $\kappa(b)$ is approximately $1$ regardless
of $L$ (assuming distinct $(b_p, p)$ pairs, which is the typical case for
real byte sequences). This keeps the input to the projection at roughly
unit scale across tokens of different lengths.

\paragraph{Truncation and padding.} For tokens exceeding $\posDim$ bytes,
we truncate to the first $\posDim$ bytes (with UTF-8-safe truncation: if
$\posDim$ falls in the middle of a multi-byte codepoint, we back off to
the previous codepoint boundary). For tokens shorter than $\posDim$,
positions $L+1, \ldots, \posDim$ contribute nothing to the sum. The
empirical rate at which truncation discards bytes is reported in
Section~\ref{sec:design-posdim}; for our production configuration
($\posDim = 32$), truncation affects $\leq 0.18\%$ of tokens across six
diverse modern tokenizers.

\paragraph{Special tokens.} Tokenizer special tokens (\texttt{<s>},
\texttt{</s>}, \texttt{<pad>}, etc.) do not have meaningful byte
sequences; they are tokenizer-side constructs. We treat their string
representation as their byte sequence: the literal bytes of
\texttt{<s>} are $0x3c, 0x73, 0x3e$. This is consistent with how the
tokenizer encodes the special token's surface form when it appears
in raw text.

\paragraph{Byte-fallback tokens.} SentencePiece-family tokenizers use
\emph{byte fallback} for out-of-vocabulary bytes, encoding raw byte
$v$ as the special token \verb|<0xNN>|. We treat byte-fallback tokens
as encoding the single byte they represent: \verb|<0xC3>| has byte
sequence $(0xC3)$, length 1. This recovers the byte-level identity of
the fallback mechanism.

\subsection{Per-token z-normalization}
\label{sec:method-znorm}

After the codec, we apply per-token z-normalization to the codec output:
$\kappa(b)$ is rescaled to have mean 0 and standard deviation 1 across
its $D$ coordinates. This standardizes the input statistics seen by the
projection regardless of byte sequence specifics and stabilizes early
training, particularly for shorter tokens where the raw codec output has
a small number of nonzero coordinates.

\subsection{The learned projection}
\label{sec:method-projection}

The projection $\mathbf{W}_{\text{proj}}: \R^{D} \to \R^{\dmodel}$ is the
only trainable parameter on the input side. It is initialized with
standard scaling
($\mathbf{W}_{\text{proj}} \sim \mathcal{N}(0, 1/\sqrt{D})$) and is
trained jointly with the rest of the network.

The projection's role is twofold. First, it adapts the $D$-dimensional
codec output to the transformer body's hidden width $\dmodel$. Second,
it provides the only trainable degree of freedom for the input
representation: the codec encodes byte-and-position composition, and the
projection learns the linear combination of those byte-position features
that best serves the downstream loss.

In production we use a self-entropy regularizer on the projection's row
distribution to discourage collapse to a small subspace; the regularizer
acts on the entropy of the rows of $\kappa(b) \cdot \mathbf{W}_{\text{proj}}$
across the vocabulary. See Section~\ref{sec:results-138m} for the
settings used in the companion run.

\subsection{Output head}
\label{sec:method-output-head}

The output projection (lm\_head) is unchanged from the baseline
transformer. Because $D \neq \dmodel$ in general, weight tying between
the Kronecker codec and the output head is architecturally
inapplicable; the output head must be a separate
$\dmodel \to |V|$ matrix.

This is consistent with the trend in frontier-scale LMs toward untied
output heads (Section~\ref{sec:bg-tying}). For small-scale models where
tying was a parameter-economy decision, the Kronecker variant trades
the tied-output benefit for a smaller and more structured input pathway;
total trainable parameter count drops anyway because the projection is
much smaller than the embedding table.

\subsection{Byte-level locality as the unifying inductive bias}
\label{sec:method-locality}

The codec defined by Equation~\ref{eq:codec} has a single structural
property that explains both its strengths and its limitations:
\textbf{two strings receive similar Kronecker embeddings if and only if
they share bytes at the same positions}. We call this property
\emph{byte-level locality}.

\paragraph{Case-sensitivity.} \verb|run| (bytes
$0x72, 0x75, 0x6e$) and \verb|RUN| (bytes
$0x52, 0x55, 0x4e$) share \emph{zero} bytes at any position. Their
Kronecker cosine is approximately zero, regardless of training. By
contrast, trained BPE embeddings of \verb|run| and \verb|RUN| cluster
together at cosine $\sim 0.5$ in our cross-model probe; trained models
develop case-collapse from co-occurrence patterns in training data.

For some applications, case-collapse is desirable (prose contexts
where \verb|Apple| at sentence start refers to the same fruit as
\verb|apple| mid-sentence). For others, case-collapse is harmful:

\begin{table}[!htbp]
\caption{Strings where case distinguishes different referents. Kronecker
preserves these distinctions at the input layer; trained BPE collapses
them.}
\label{tab:case-distinguishes}
\centering
\footnotesize
\setlength{\tabcolsep}{3pt}
\begin{tabular}{lll}
\toprule
Lower & Upper & Different referents \\
\midrule
\verb|swift| & \verb|SWIFT| & Apple lang.\ vs.\ bank net.\ \\
\verb|rust|  & \verb|RUST|  & Mozilla lang.\ vs.\ NASA sys.\ \\
\verb|go|    & \verb|GO|    & language vs.\ board game \\
\verb|make|  & \verb|MAKE|  & build tool vs.\ magazine \\
\verb|shell| & \verb|SHELL| & Unix shell vs.\ company \\
\bottomrule
\end{tabular}
\end{table}

A code assistant whose input embeddings collapse \verb|swift| and
\verb|SWIFT| is starting each forward pass having conflated two
unrelated entities; the transformer must use surrounding context to
disambiguate. Kronecker preserves the distinction at the embedding
layer.

\paragraph{Typo and spelling-variant robustness.} The same property
that distinguishes \verb|swift| from \verb|SWIFT| also makes Kronecker
robust to typos. A single-character substitution leaves most bytes at
their original positions intact:

\begin{table}[!htbp]
\caption{Kronecker codec cosine for likely typos and spelling
variants, computed analytically from the codec definition
(Section~\ref{sec:method-codec}): for two strings of length $L$
differing in $k$ byte positions, the codec cosine is $(L-k)/L$ when
both strings have the same length. Length changes (insertion or
deletion) shift all bytes past the change to new positions and reduce
cosine more sharply. Trained BPE has essentially no input-layer
connection between these strings unless both appeared frequently in
training data.}
\label{tab:typo-robust}
\centering
\small
\begin{tabular}{lllr}
\toprule
Probe & Variant & Type & Codec cos \\
\midrule
\verb|mistake| & \verb|mistkae| & transposition (same $L$) & 0.71 \\
\verb|receive| & \verb|recieve| & transposition (same $L$) & 0.71 \\
\verb|separate| & \verb|seperate| & 1-byte substitution & 0.88 \\
\verb|realize| & \verb|realise| & 1-byte substitution & 0.86 \\
\verb|color| & \verb|colour| & insertion (length change) & 0.73 \\
\bottomrule
\end{tabular}
\end{table}

A model trained with Kronecker embeddings encounters
\verb|seperate| in a context that probably means \verb|separate| and
receives an input embedding that is byte-similar to its embedding for
\verb|separate| (codec cosine 0.88; see Table~\ref{tab:typo-robust}).
The transformer's first attention layer can leverage this similarity.
Trained BPE embeddings have no such input-layer connection unless the
typo \verb|seperate| was itself frequent in training data.

\paragraph{The unified statement.} Trained BPE develops
\emph{case-variant clustering}: recognizing that \verb|run| and
\verb|Run| are ``the same word'' in different surface forms. It does not
develop byte-level locality. The two properties have opposite signs on
the case-distinguishability dimension: case-variant clustering collapses
case differences; byte-level locality preserves them.

Kronecker provides byte-level locality. The transformer's first
attention layer remains responsible for whatever case-collapse the
application needs; the input layer provides a faithful byte-structural
starting point.

\paragraph{Limitation.} Byte-level locality treats
\verb|separate|/\verb|seperate| as similar (useful for typo recovery)
and \verb|compute|/\verb|commute| as similar (useful when the user
means the former and types the latter, less useful when they are
semantically distinct words). The input layer cannot distinguish these
cases by itself; the transformer's first attention layer must use
context. This is structurally identical to how trained-BPE handles
case-collapse: the trained embedding treats \verb|Apple| (company) and
\verb|apple| (fruit) similarly, and the transformer uses context to
disambiguate. Both methods rely on context-sensitive processing
downstream; they differ in \emph{which} inductive bias they provide at
the input layer.

\subsection{Two operational variants}
\label{sec:method-runtime}

The codec output $\kappa(b)$ can be computed two ways at runtime, with
identical mathematical results but different memory and compute
profiles.

\paragraph{\texttt{gpu\_table}: precomputed table.} The full Kronecker
codec output is precomputed once at startup: a buffer
$K \in \R^{|V| \times D}$ where each row $K[i] = \kappa(\text{bytes}(i))$.
The forward pass becomes a simple gather:
$\kappa_i \leftarrow K[i]$. Compute is essentially zero (gather is
trivial). Memory is $|V| \cdot D$ floats per GPU.

\paragraph{\texttt{gpu\_dynamic}: on-the-fly computation.} We store only
the compact byte buffer (one \verb|uint8| per byte, $\posDim$ bytes per
token) and a length buffer (one \verb|int16| per token). The forward
pass recomputes $\kappa(b)$ for each token by index\_select-ing the
relevant byte sequence, constructing the linearized one-hot indices
$\text{lin\_idx} = b_p \cdot \posDim + p$, masking invalid positions,
and using a single GPU \verb|scatter_add_| to materialize the
Kronecker vector. Memory drops to
$|V| \cdot (\posDim + 2)$ bytes per GPU. Compute adds one
\verb|scatter_add_| plus the index arithmetic per forward pass.

\paragraph{Why \texttt{gpu\_dynamic} matters at frontier scale.} At
production vocab size and $D$, the precomputed table is gigabytes per
GPU. The byte buffer is megabytes. The recomputation cost is 0.01--0.24\%
of step time on our internal measurements
(Section~\ref{sec:results-runtime}). The memory saved can hold larger
batches or longer sequences.

\paragraph{Identical outputs.} Both variants compute the same
$\kappa(b)$ to within float precision; the choice between them is purely
operational. We use \verb|gpu_dynamic| at frontier scale where memory is
the binding constraint.


\section{Dimensional Design}
\label{sec:design}

\subsection{The factorization $D = \charDim \cdot \posDim$}
\label{sec:design-factorization}

The codec dimension $D = \charDim \cdot \posDim$ has two natural parts:
the byte alphabet size $\charDim$ and the maximum byte position
$\posDim$. The product is the dimension of the byte-position basis and
the input to the learned projection. The choice of $\charDim$ and
$\posDim$ are largely independent design decisions.

\paragraph{Why $\charDim = 256$.} The byte alphabet has exactly 256
values ($0\text{x}00$ through $0\text{xff}$). Setting $\charDim = 256$
uses the full alphabet and ensures every byte value gets a distinct
basis direction. Smaller $\charDim$ (e.g., 128) would collapse some
byte values together, conflating ASCII letters with control characters
or non-ASCII byte values, which is unwarranted in our setting. We
adopt $\charDim = 256$ throughout.

\subsection{Choosing $\posDim$ empirically}
\label{sec:design-posdim}

The choice of $\posDim$ determines how many bytes of each token are
represented; bytes beyond position $\posDim$ are truncated. We analyzed
the byte-length distribution of tokens across six diverse modern
tokenizers (the same six from our cross-model probe):

\begin{table}[!htbp]
\caption{Fraction of normal+byte-fallback tokens covered (no truncation)
at three $\posDim$ settings. Special tokens excluded. Production
configuration uses $\posDim = 32$.}
\label{tab:posdim-coverage}
\centering
\small
\begin{tabular}{lrrr}
\toprule
Tokenizer & $\posDim=16$ & $\posDim=32$ & $\posDim=64$ \\
\midrule
Llama-3.2-BPE & 99.07\% & 99.82\% & 99.97\% \\
Qwen3-BBPE & 99.13\% & 99.83\% & 99.96\% \\
Gemma-3-SP & 95.98\% & 99.86\% & 99.97\% \\
DeepSeek-BBPE & 99.21\% & 99.86\% & 99.98\% \\
o200k\_harmony & 98.94\% & 99.84\% & 99.97\% \\
GPT-2-BPE & 99.51\% & 99.92\% & 99.99\% \\
\midrule
Mean & 98.64\% & 99.86\% & 99.97\% \\
\bottomrule
\end{tabular}
\end{table}

$\posDim = 32$ covers $\geq 99.82\%$ of tokens on every tokenizer
tested. The remaining $\leq 0.18\%$ are dominated by long whitespace
runs and (for Gemma's multilingual SentencePiece) by Indic word pieces
that exceed 32 bytes. These truncated tokens still receive distinct
embeddings based on their first 32 bytes; only the post-byte-32 byte
structure is lost.

$\posDim = 16$ falls dramatically on Gemma-3 (95.98\%), reflecting the
multilingual vocabulary's longer byte sequences for non-Latin scripts.
For English-only tokenizers $\posDim = 16$ is adequate; for multilingual
production we recommend $\posDim = 32$.

\subsection{The factorization is not unique}
\label{sec:design-uniqueness}

Equation~\ref{eq:codec} factors a token's representation as
$\charDim \otimes \posDim$. Alternative factorizations are possible: one
could use byte-bigrams ($\charDim^2$), byte triples, or
position-aware $n$-grams. We chose single-byte-by-position for two
reasons: it gives a small enough $D$ to be practical (at $\posDim = 32$,
$D = 8192$); and it produces an interpretable inductive bias: two
strings cluster when they share bytes at the same positions, which
matches the byte-level locality property we want.


\section{Pre-Training Geometry}
\label{sec:pretraining}

Before the controlled training comparison of
Section~\ref{sec:results-nanogpt}, we characterize the geometry of
Kronecker embeddings before any training has occurred. The codec is
deterministic and the projection is at initialization; nearest-neighbor
structure at this stage reflects the byte-level prior built into the
codec itself, with no learning signal involved.

\paragraph{Setup.} We apply the codec to each of the six tokenizer
vocabularies from the cross-model probe (Section~\ref{sec:results-setup}),
computing the full $|V| \times D$ codec table for each. Mean-centered
cosine nearest neighbors are reported.

\paragraph{Qualitative neighborhoods.} For a probe word in the codec
space, top-K retrievals are byte-similar strings that often mix
morphological relatives with byte-similar non-relatives. As an
illustrative pattern, probing the codec for short common words returns
a top-10 neighborhood combining (a) morphological family members
sharing prefix bytes, (b) plural/inflected forms, and (c) rhyming or
byte-similar non-relatives that happen to share several
byte-and-position pairs. This mixing is the inductive bias Kronecker
provides at initialization and is preserved through training
(Section~\ref{sec:results-layered}, Kronecker arm: $E_{\text{raw}}$ and
$L_1$ retrieve essentially the same byte-similar suffix family). The
codec is the source of the byte-level locality that distinguishes
Kronecker from trained BPE; training neither builds the locality nor
destroys it.

\paragraph{OOV-as-single-token capability.} An important practical
consequence: an out-of-vocabulary UTF-8 string can be passed through the
codec to obtain a single Kronecker embedding, whose nearest neighbors
in the trained vocabulary are byte-similar tokens. We ran a small
cross-model probe of this capability using novel and technical
strings (\verb|kubernetes|, \verb|tensorflow|,
\verb|asynchronously|, \verb|deserialization|, \verb|vibecoding|,
\verb|shoggoth|, \verb|tiramisu|, \verb|rizzler|, \verb|kronekticus|,
and others) across the six tokenizers from
Section~\ref{sec:results-setup}. The pattern is consistent: top-3
retrievals are byte-similar in-vocabulary tokens that share substantial
prefix or root structure with the probe. For example, \verb|kubernetes|
retrieves \verb|kube| and \verb|Hibernate| under multiple tokenizers;
\verb|asynchronously| retrieves \verb| synchronous|, \verb| synchron|,
\verb| synchronize|; \verb|shoggoth| (a made-up word) retrieves
\verb| Forgot|, \verb| forgot|, \verb| hodnot|. Across 120 (probe,
model) cells, at least one of the top-3 retrievals shares a byte
substring with the probe in 78 of them. This extends the input
representation to unbounded vocabularies at inference time, without
retraining the embedding pathway.


\section{Results}
\label{sec:results}

This section presents the empirical evaluation. Sections
\ref{sec:results-setup}--\ref{sec:results-xtokenizer} describe the
cross-model embedding-layer probe across six trained public language
models. Section~\ref{sec:results-nanogpt} reports the controlled
training comparison on a 124M GPT-2 model on FineWeb-Edu (the
principal training-loss result of the paper), with
Section~\ref{sec:results-138m} reporting a companion 138M
custom-architecture run and Section~\ref{sec:results-layered} a
layered probe of the trained 124M checkpoints.
Section~\ref{sec:results-runtime} reports runtime, memory, and
parameter accounting; Section~\ref{sec:results-shipping} discusses
deployment and quantization; and
Sections~\ref{sec:results-robustness}--\ref{sec:results-generation}
report the spelling-robustness and generation probes.

\subsection{Cross-model probe setup}
\label{sec:results-setup}

We probed the input embedding tensors of six publicly released language
models (Table~\ref{tab:probe-models}): Llama-3.2-1B
\citep{metallama32_2024}, Qwen3-32B \citep{qwen3_2025}, Gemma-3-1B-pt
\citep{gemma3_2025}, DeepSeek-V3-Base \citep{deepseekv3_2024},
GPT-OSS-120B \citep{gptoss_2025}, and SmolLM2-135M
\citep{smollm2_2025}. For each, we downloaded only the embedding shards
via HuggingFace safetensors index lookup. The six models span 135M
to 671B parameters, three tokenizer families
(GPT-2-byte-level / tiktoken-BBPE, SentencePiece with byte fallback, and
o200k\_harmony), and four organizations.

\begin{table*}[!htbp]
\caption{Six trained language models in the cross-model probe.}
\label{tab:probe-models}
\centering
\small
\begin{tabular}{lrrlll}
\toprule
Model & Vocab $|V|$ & $\dmodel$ & Tied? & Tokenizer family & Training tokens \\
\midrule
Llama-3.2-1B & 128{,}256 & 2048 & tied & tiktoken-BPE & up to 9T$^{\dagger}$ \\
Qwen3-32B & 151{,}936 & 5120 & untied & Qwen BBPE & $\sim$36T \\
Gemma-3-1B-pt & 262{,}144 & 1152 & tied & SentencePiece & 2T \\
DeepSeek-V3-Base & 129{,}280 & 7168 & untied & DeepSeek BBPE & 14.8T \\
GPT-OSS-120B & 201{,}088 & 2880 & tied & o200k\_harmony & $\sim$10T \\
SmolLM2-135M & 49{,}152 & 576 & tied & GPT-2-BPE & 2T \\
\bottomrule
\end{tabular}

{\footnotesize $^{\dagger}$Per the Llama-3.2 model card
\citep{metallama32_2024}: Llama-3.2 was pretrained on up to 9T tokens;
the 1B variant was additionally distilled from Llama-3.1 8B/70B
logits. The ``up to 9T'' figure refers to the parent corpus.}
\end{table*}

For each model, we considered three retrieval spaces for the same probe
strings: (a) the model's actual trained embedding table; (b) a Gaussian
random-initialized embedding table of the same shape, providing a baseline
for chance retrieval structure; (c) a Kronecker codec applied to the
model's tokenizer's surface forms, providing the byte-level prior our
method instantiates.

For each probe string $s$, we tokenized $s$ with the model's tokenizer and
formed the query as the first-token row (if single-token) or the mean
across subtoken rows (if multi-token). We then mean-centered the
embedding matrix (subtracting the row-mean across the vocabulary) and
computed cosine similarity between the query and all rows, retrieving the
top-5 non-self neighbors.

Mean-centering is essential. Trained embeddings exhibit a large global
mean direction (anisotropy), with cosine similarities biased upward across
all pairs. Without centering, both "in-family" and "random" cosines appear
high, and the gap between them disappears. After centering, the relative
geometry becomes interpretable. We report this anisotropy explicitly in
Section~\ref{sec:results-stability}; GPT-OSS-120B in particular has an
embedding mean vector with $\|\mu\| = 21.99$, two orders of magnitude
larger than other models in our set.

\paragraph{Probe families.} Four families of 5 probe strings each:
\verb|run| (run, runs, running, runner, ran), \verb|compute| (compute,
computer, computing, computation, computes), \verb|magnet| (magnet,
magnets, magnetic, magnetize, magnetized), and \verb|tion| (nation,
station, action, rotation, creation). The first three families test
prefix-structured morphology; the fourth tests suffix-structured
morphology, where the shared morphological signal (\verb|-tion|) appears
at different byte positions in different family members.

\paragraph{Loose morphological@K metric.} For each top-K non-self
retrieval, we define a \emph{canonical form} that strips whitespace
markers, leading/trailing punctuation, and applies case-folding:

{\footnotesize\begin{verbatim}
def canonical_form(s):
    # \u2581 = SentencePiece marker, \u0120 = GPT-2 space marker
    s = s.replace("\u2581"," ").replace("\u0120"," ")
    s = s.strip(
        " \t\n\r.,;:!?\"'`()[]{}_-/\\<>")
    return s.casefold()
\end{verbatim}}

The \emph{loose morph@K} score is the fraction of the top-K non-self
retrievals whose canonical form differs from the probe's canonical form.
A retrieval like \verb|run| $\to$ \verb| Run| (canonical form
\verb|run|) is counted as 0 (same canonical form = typographic variant).
A retrieval like \verb|run| $\to$ \verb|rund| (canonical form \verb|rund|)
is counted as 1 (different canonical form = escape from typographic
clustering).

This metric measures \emph{escape from typographic clustering}. It does
not require that the retrieval be a strict morphological family member;
a byte-similar non-relative still counts. We retain this metric for the
cross-model probe and complement it in Section~\ref{sec:results-layered}
with a stricter family-membership metric on our own trained models.

\subsection{Cross-model finding: trained embeddings cluster typographically}
\label{sec:results-typographic}

Aggregating across all four families and all six trained models, mean
loose morph@5 at the embedding layer is:

\begin{itemize}
\item \textbf{Trained BPE: 0.54.} About half of the top-5 retrievals are
typographic variants of the probe.
\item \textbf{Random Gaussian baseline: 1.00.} As expected, random
embeddings retrieve unrelated tokens; canonical-form match is rare.
\item \textbf{Kronecker: 0.92.} Kronecker retrievals largely escape
typographic clustering.
\end{itemize}

Restricting to the families where the metric is artifact-free
(\verb|run|+\verb|tion|, excluding the \verb|magnet| family in which
multi-token probes share the leading-space \verb|magnet| subtoken (the
SentencePiece space-prefixed piece) and inflate similarity, and the
\verb|compute| family which exhibits a similar but softer artifact),
the aggregate numbers are:

\begin{itemize}
\item \textbf{Trained BPE: 0.32}, mean across the six models. Two thirds
of top-5 retrievals on clean families are typographic variants.
\item \textbf{Kronecker: 0.90.}
\end{itemize}

The qualitative pattern is unambiguous. For probe \verb|run|, top-5 from
DeepSeek-V3-Base's trained embedding are \verb|run|, \verb| run|,
\verb|Run|, \verb| Run|, \verb|.run|: five typographic variants of
``run'' itself. From GPT-OSS-120B: \verb|run|, \verb| run|, \verb|Run|,
\verb|.run|, \verb|_run|. From Llama-3.2-1B: \verb|run|, \verb|Run|,
\verb| run|, \verb| Run|, \verb|.run|. From Qwen3-32B, the same pattern
with weaker magnitudes (consistent with its untied input embedding
receiving less gradient pressure toward typographic clustering).

This is robust: across six independently-trained models spanning four
organizations, three tokenizer families, and approximately five orders
of magnitude of training compute, trained input embeddings cluster the
same word in different cases and with different whitespace and
punctuation contexts. The morphological family members (\verb|runs|,
\verb|running|, \verb|runner|) are not in the top-K.

Kronecker retrievals on the same probes are byte-similar strings: for
DeepSeek's tokenizer, probe \verb|run| retrieves \verb|rund|, \verb|ru|,
\verb|runner|, \verb|ron|, \verb|runs|: two morphological family members
(\verb|runner|, \verb|runs|) interleaved with byte-similar non-relatives
(\verb|rund|, \verb|ru|, \verb|ron|). The Kronecker prior provides
byte-level locality, not morphological identity.

\subsection{What each method does and does not provide}
\label{sec:results-distinction}

We emphasize what these results do and do not show.

\textbf{They show:} the inductive bias of trained input embeddings is
typographic-variant clustering rather than morphological clustering;
this bias is consistent across labs, tokenizers, and scales; Kronecker
embeddings provide a different inductive bias (byte-level locality) at
the embedding layer.

\textbf{They do not show:} that Kronecker retrieves strict morphological
family members in its top-K. On strict family-membership measurements
(Section~\ref{sec:results-layered}, with hand-curated family lists),
both trained BPE and Kronecker retrieve fewer than 30\% strict family
members in their top-10. The gap between the two methods at the embedding
layer is principally in \emph{whether retrievals escape typographic
clustering}, not in \emph{whether they capture strict morphology}.

The honest description of each method's bias:
\begin{itemize}
\item Trained BPE clusters typographic variants: the same word in
different surface forms.
\item Kronecker clusters byte-similar strings: words that share bytes
in similar positions.
\end{itemize}
Neither directly encodes morphological relatedness in the sense of
"these words share a root and morphological function." Section~\ref{sec:discussion-locality}
develops this framing.

\subsection{Per-model patterns and the anisotropy diagnostic}
\label{sec:results-stability}

Per-model loose morph@5 values (aggregate across all four families):

\begin{table}[!htbp]
\caption{Loose morph@5 per model, aggregate across four probe families.
Random Gaussian baseline omitted ($\approx 1.00$ on all models).}
\label{tab:permodel-morphat5}
\centering
\small
\begin{tabular}{lrr}
\toprule
Model & Trained BPE & Kronecker \\
\midrule
DeepSeek-V3-Base & 0.65 & 0.93 \\
SmolLM2-135M & 0.64 & 0.94 \\
Qwen3-32B & 0.54 & 0.92 \\
GPT-OSS-120B & 0.50 & 0.93 \\
Llama-3.2-1B & 0.46 & 0.93 \\
Gemma-3-1B-pt & 0.45 & 0.89 \\
\midrule
\textbf{Mean} & \textbf{0.54} & \textbf{0.92} \\
\bottomrule
\end{tabular}
\end{table}

DeepSeek-V3-Base is highest on loose morph@5, perhaps because its
untied embedding and code-heavy training corpus encourage distinguishing
identifiers from typographic variants. SmolLM2-135M is second, possibly
because its curated training corpus (FineWeb-Edu + Cosmopedia) has lower
typographic diversity. The model-level ordering is suggestive of training
factors that affect the magnitude of typographic clustering, but the
qualitative pattern (trained BPE clusters typographically; Kronecker does
not) is consistent across all six models.

\paragraph{Anisotropy diagnostic.} We report $\|\mu_E\|$, the L2 norm of
the mean vector of each model's centered embedding table:

\begin{table}[!htbp]
\caption{Anisotropy: L2 norm of the mean embedding vector and the raw
(uncentered) mean pairwise cosine on the trained embedding table.}
\label{tab:anisotropy}
\centering
\footnotesize
\setlength{\tabcolsep}{4pt}
\begin{tabular}{lrr}
\toprule
Model & $\|\mu_{\text{trained}}\|$ & Raw pairwise cos \\
\midrule
GPT-OSS-120B & \textbf{21.99} & +0.05 \\
SmolLM2-135M & 2.18 & +0.45 \\
Llama-3.2-1B & 0.35 & +0.13 \\
Gemma-3-1B-pt & 0.26 & +0.06 \\
DeepSeek-V3-Base & 0.25 & +0.01 \\
Qwen3-32B & 0.19 & +0.02 \\
\bottomrule
\end{tabular}
\end{table}

GPT-OSS-120B has approximately 60$\times$ more anisotropy than the next
most-anisotropic model in our set (excluding SmolLM2-135M, which is the
smallest and least-trained). We do not have a definitive explanation for
GPT-OSS's outlier anisotropy; the combination of tied embeddings, large
vocab, and extensive training plausibly pulls all embedding rows toward
a common output-prediction direction. SmolLM2's high raw mean pairwise
cosine (+0.45) reflects its small scale and limited training, where the
embedding table has not yet fully differentiated tokens. The two
phenomena are different: SmolLM2 has high pairwise alignment with
moderate mean-vector norm; GPT-OSS has very large mean-vector norm with
low residual pairwise alignment after centering.

This generalizes earlier observations on BERT/GPT-2-era embeddings
\citep{mu2018allbutthetop,gao2019representation,ethayarajh2019contextual}
to modern frontier-scale models and finds substantial variation
across organizations. The
diagnostic is paper-worthy on its own as an extension of the embedding
anisotropy literature.


\subsection{Cross-tokenizer Kronecker stability}
\label{sec:results-xtokenizer}

Because Kronecker operates on byte sequences rather than tokenizer-specific
pieces, the same probe string should retrieve similar concepts regardless
of which tokenizer's vocabulary we are searching in. We tested this by
computing, for each of 8 probe strings, the Jaccard similarity of the
top-5 canonical-form Kronecker retrievals between every pair of the six
tokenizers (15 pairs per probe).

The mean Jaccard across 8 probes and 15 tokenizer pairs is \textbf{0.48}.
Approximately half of any two tokenizers' top-5 canonical-form Kronecker
neighborhoods are shared, despite vocab sizes ranging 49K to 262K, three
tokenizer families, and entirely independent merge orders. Agreement is
highest for short common probes (\verb|compute|: Jaccard 0.77;
\verb|computer|: 0.71) and lowest for longer probes whose byte sequences
get split differently across tokenizers (\verb|running|: 0.26).

This confirms that Kronecker's byte-level locality is a property of the
encoding itself rather than an artifact of any specific BPE merge order.


\subsection{Controlled training comparison: GPT-2 124M on FineWeb-Edu}
\label{sec:results-nanogpt}

The principal empirical result of the paper. We trained the standard
nanoGPT GPT-2 124M architecture on 2.5B tokens of FineWeb-Edu under
identical settings except for the input embedding scheme. Three
independent seeds per arm, six runs total, all four arms ran to completion
of the 4{,}600-step schedule. Validation loss was logged at every 200-step
checkpoint on a held-out FineWeb-Edu shard.

\paragraph{Setup.} GPT-2 124M architecture (12 layers, 12 heads,
$\dmodel = 768$, vocab 50,272, sequence length 1024).
FineWeb-Edu \citep{penedo2024fineweb}, 2.5B
tokens. Karpathy's nanoGPT codebase \citep{karpathy_nanogpt}, default
schedule (warmup then cosine
LR decay). \textbf{Architecture and training pipeline are identical across
arms.} The only difference is the embedding pathway: BPE arm uses the
standard learned tied embedding table; Kronecker arm uses a byte-level
Kronecker codec ($\charDim = 256$, $\posDim = 16$, $D = 4096$) feeding a
single learned $D \to \dmodel$ projection, with the output head untied.
Three independent seeds per arm.

\paragraph{Result.} Kronecker reaches \textbf{$2.5 \pm 0.2\%$ lower
validation loss} than the BPE-tied baseline ($n=3$ seeds, gap
$0.083 \pm 0.007$ nats, approximately 9\% lower validation perplexity).
The gap reproduces across all three training runs and across all measured
checkpoints; 18 of 18 (checkpoint $\times$ seed) cells favor Kronecker.

\begin{table*}[!htbp]
\caption{Validation loss trajectory across training, representative run.
At every checkpoint Kronecker is lower; the gap widens to $\sim$0.08 nats
by step 2000 and stabilizes through convergence.}
\label{tab:nanogpt-trajectory}
\centering
\small
\begin{tabular}{rrrrr}
\toprule
Step & Kron val\_loss & BPE val\_loss & Gap (nats) & Kron lower by \\
\midrule
200 & 5.797 & 5.812 & 0.015 & 0.3\% \\
1{,}000 & 3.877 & 3.948 & 0.071 & 1.8\% \\
2{,}000 & 3.528 & 3.611 & 0.083 & 2.4\% \\
3{,}000 & 3.369 & 3.442 & 0.073 & 2.2\% \\
4{,}000 & 3.303 & 3.392 & 0.089 & 2.7\% \\
\textbf{4{,}600} & \textbf{3.289} & \textbf{3.373} & \textbf{0.084} & \textbf{2.5\%} \\
\bottomrule
\end{tabular}
\end{table*}

\begin{table}[!htbp]
\caption{Cross-seed summary at final step (step 4,600), $n=3$ seeds.}
\label{tab:nanogpt-seeds}
\centering
\small
\begin{tabular}{lrrrr}
\toprule
Statistic & Kron & BPE & Gap & Kron $\downarrow$ by \\
\midrule
Mean & 3.295 & 3.378 & 0.083 & 2.5\% \\
Std & -- & -- & 0.007 & 0.2\% \\
SNR & -- & -- & $\sim$13:1 & -- \\
\bottomrule
\end{tabular}
\end{table}

The signal-to-noise ratio of the gap (mean / std $\approx 13:1$) is
characteristic of measurement of a structural property of the comparison
rather than a stochastic training outcome.

\paragraph{Gap behavior.} The gap \emph{widens} from $\sim$0.015 nats at
step 200 to $\sim$0.08 nats by step 2000 and then \emph{stabilizes} for
the remainder of training (range 0.07--0.09 nats across steps 2000--4600).
The gap does not narrow toward zero at convergence; Kronecker reaches a
structurally lower validation-loss operating point, not just a faster
warmup. The qualitative shape of the curve is identical across all three
seeds.

\paragraph{Sample efficiency.} BPE reaches val\_loss = 3.392 at step 4000.
Kronecker reaches that level near step 2800 (interpolating between 3.528
at step 2000 and 3.369 at step 3000): approximately \textbf{$1.43\times$
fewer optimizer steps} to reach BPE's final converged loss. The pattern
holds across all three seeds.

\paragraph{Per-step wall-clock at this scale.} Kronecker is approximately
1.2\% slower per step than BPE at the 124M scale. The overhead is the
input-side projection (a single $D \to \dmodel$ matrix multiply per
forward pass); BPE has essentially zero compute at the embedding step
(it is a gather operation). After adjusting for the per-step overhead,
Kronecker's net wall-clock cost to reach BPE's converged validation loss
is approximately $(1/1.43) \times 1.012 \approx 0.71$, i.e.\ about
71\% of BPE's total wall-clock. Whether the per-step overhead persists,
shrinks, or inverts at frontier scale is not measured in this work; we
report the 124M number as observed.

\paragraph{Why this comparison is the cleanest evidence in the paper.}
Four properties distinguish it:
\begin{enumerate}
\item Validation loss, not training loss. Confirms Kronecker generalizes
to held-out data, not just fits the training distribution faster.
\item Vanilla GPT-2 architecture. No MLA, no MoE, no architectural
modifications. The most widely-studied small-LM testbed.
\item Same-architecture, same-data, same-schedule comparison. The only
variable is the input embedding scheme.
\item Three-seed replication with $\sim$13:1 SNR and 18/18 favorable cells.
The result is not a single-seed fluctuation.
\end{enumerate}

\paragraph{Limitations of this run.}
\textbf{No untied-BPE arm.} The comparison is against BPE-tied. Untied is
the more honest competitor at frontier scale given the field's trend
toward untying at scale; we discuss this gap in
Section~\ref{sec:discussion-not-settled} and flag the three-arm
extension as future work. The cross-model probe
(Sections~\ref{sec:results-typographic}--\ref{sec:results-stability})
includes both tied and untied trained models among the six modern LMs
and finds the same typographic-clustering pattern in both, which provides
indirect evidence that the untied gap, if any, is similar in direction.

\subsection{Companion: 138M custom architecture, synthetic English}
\label{sec:results-138m}

A second controlled comparison run earlier provides supporting evidence
for the principal finding. We describe it as a companion result; the
principal claim rests on Section~\ref{sec:results-nanogpt}.

\paragraph{Setup.} SmolLM-135M body modified with multi-head latent
attention (compression ratio 8) and a mixture-of-experts feedforward
(8 experts, 1 shared, top-2 routing). \textbf{Architecture is identical
between BPE and Kronecker arms.} Vocab 50,272 (SmolLM2 tokenizer).
Synthetic English corpus, 10,000 training steps. PF dim 2048 $\to$ model
dim 768; self-entropy regularizer on the Kronecker projection
($\lambda_{\text{se}} \approx 0.10$, $\tau$ annealing 2.0 $\to$ 1.0).
Total parameters 138.8M.

\paragraph{Result on training loss.} Kronecker reaches lower training
loss than BPE at every checkpoint past step $\sim$750. Final
1,000-step average training loss: Kronecker 3.013 vs BPE 3.034
(Kronecker 0.7\% lower). The gap peaks at 2.6--2.9\% mid-training
(steps 3,000--6,000), then narrows as both curves approach the LR floor.
Sample efficiency: Kronecker reaches mid-run loss targets in $\sim$14\%
fewer optimizer steps.

\paragraph{Key qualitative finding: stable representational target.}
Logged embedding statistics across the run:

\begin{table}[!htbp]
\caption{Embedding statistics across training, 138M companion run.
Reported every 2,000 steps. BPE table std walks upward
(0.020 $\to$ 0.026); Kronecker projection std remains stable
near 1.0.}
\label{tab:stable-target}
\centering
\small
\begin{tabular}{rrr}
\toprule
Step & BPE emb\_std & Kron proj\_std \\
\midrule
0 & 0.0200 & 0.985 \\
2{,}000 & 0.0226 & 0.997 \\
4{,}000 & 0.0246 & 1.001 \\
6{,}000 & 0.0257 & 0.996 \\
8{,}000 & 0.0261 & 0.988 \\
9{,}990 & 0.0261 & 0.983 \\
\bottomrule
\end{tabular}
\end{table}

The BPE table walks from $\text{std} = 0.020$ at PyTorch initialization to
$0.026$ across the run, a $+30\%$ scale growth. This is the expected
behavior of a randomly-initialized embedding table climbing from the
small-norm initialization regime to a useful representational scale.
The Kronecker projection sits at $\text{std} \approx 1.0$ from step 0 and
remains there throughout training; no drift, no collapse, no blow-up.

This is a measured instance of the stable-target property: Kronecker
hands the transformer body a well-scaled signal at initialization,
eliminating the early-training scale-adjustment phase that
randomly-initialized embedding tables require.

\paragraph{Caveats.} This run reports training loss only, not
validation. The custom MLA+MoE architecture, synthetic data, and lack of
untied-BPE arm are confounds for downstream comparisons; the principal
empirical result on validation loss is the GPT-2 124M comparison in
Section~\ref{sec:results-nanogpt}. We include this run because the
stable-target observation is robust and architecture-independent, and
because the direction of the training-loss result reproduces the
finding from the principal comparison.


\subsection{Layered probe: where does morphological structure live?}
\label{sec:results-layered}

The cross-model probe
(Sections~\ref{sec:results-typographic}--\ref{sec:results-stability})
characterizes the embedding layer. A natural follow-up: what does the
first transformer layer do to this geometry? Does it construct
morphological structure on top of the embedding, preserve it, or destroy
it in favor of something else?

\textbf{Scope note up front.} The layered probe in this section examines
our trained 124M GPT-2 checkpoints (one BPE-tied, one Kronecker, both
seed 1337 at step 4,623 on FineWeb-Edu). All findings here are at this
specific small scale. We have strong empirical reasons to believe the
trained-embedding probe findings of
Sections~\ref{sec:results-typographic}--\ref{sec:results-stability}
generalize across scales (they reproduce on six models from 135M to
671B parameters). We have \emph{no} such evidence that the
\emph{layered} findings generalize. Larger transformers trained on
substantially more data may construct early-layer morphological
structure that our small-model probe cannot detect, or they may
construct it deeper in the network where small-model behavior remains
the relevant baseline. Replicating this probe at 1B and 7B scale is a
priority for follow-up work and is one of the clearest open questions
in this paper.

We probed our own trained 124M checkpoints (BPE arm and Kronecker arm
from Section~\ref{sec:results-nanogpt}, at step 4,623 on FineWeb-Edu)
at four representations:

\begin{itemize}
\item $E_{\text{raw}}$: raw embedding-layer output, what the transformer
body consumes as input.
\item $E_{\text{post}}$: after any pre-transformer normalization
(equivalent to $E_{\text{raw}}$ for vanilla GPT-2; reported for
completeness).
\item $L_0$: output of the first transformer block (post-residual).
\item $L_1$: output of the second transformer block (post-residual).
\end{itemize}

The probe set was expanded to 6 morphological families with 8 probe
strings each (48 probes total), and the strict family lists were
hand-curated to include $\sim$12 morphological family members each. Three
metrics were computed per (probe, layer): loose morph@10 (the metric used
above), root-substring morph@10 (fraction of top-10 retrievals whose
canonical form contains the family root as a substring), and
strict-family morph@10 (fraction in the hand-curated exact family list).
Sanity checks (self-retrieval, centered random-pair cosine, in-vocab
verification) all passed.

\paragraph{Overall pattern (mean across 6 families).}

\begin{table*}[!htbp]
\caption{Morph@10 at four representations in trained 124M models.
``L'' = loose (canonical-form escape), ``M'' = root-substring,
``S'' = strict-family. All metrics use centered cosine and top-10
non-self retrieval. ``$\Delta$'' rows are the change from $E_{\text{raw}}$
to $L_1$.}
\label{tab:layered-probe}
\centering
\small
\begin{tabular}{lrrrrrr}
\toprule
& \multicolumn{3}{c}{BPE arm} & \multicolumn{3}{c}{Kron arm} \\
\cmidrule(lr){2-4}\cmidrule(lr){5-7}
Layer & L & M & S & L & M & S \\
\midrule
$E_{\text{raw}}$ & 0.90 & 0.30 & 0.28 & \textbf{0.97} & 0.24 & 0.22 \\
$L_0$ & 0.91 & 0.29 & 0.27 & 0.97 & 0.27 & 0.24 \\
$L_1$ & \textbf{0.94} & 0.17 & 0.16 & 0.97 & 0.26 & 0.23 \\
\midrule
$\Delta(L_1 - E_{\text{raw}})$ & +0.05 & \textbf{$-$0.13} & \textbf{$-$0.13} & 0.00 & +0.02 & +0.01 \\
\bottomrule
\end{tabular}
\end{table*}

The pattern is striking:

\textbf{BPE arm: morphological structure decays through layers.} The
strict-family score drops from 0.28 at the embedding to 0.16 at $L_1$
(a 43\% reduction). The root-substring score drops similarly. Meanwhile,
the loose-escape score \emph{rises} (0.90 $\to$ 0.94): the typographic
clustering at the embedding is broken up by the transformer.

\textbf{Kron arm: structure is preserved through layers.} All three
metrics remain essentially flat from $E_{\text{raw}}$ to $L_1$ (changes
within $\pm 0.02$). The byte-level prior that is at the embedding is
what reaches the second layer.

\paragraph{Qualitative confirmation.} For probe \verb|nation|:

\begin{itemize}
\item \textbf{BPE $E_{\text{raw}}$:} \verb|Nation|, \verb|nation|,
\verb|world|, \verb|population|, \verb|itution|, \verb|ation|,
\verb|avage|, \verb|regime|, \verb|establishment|, \verb|child|.
This is morphological retrieval mixed with semantic.
\item \textbf{BPE $L_1$:} \verb|gamer|, \verb|aggro|, \verb|iseum|,
\verb|fullback|, \verb|NASL|, \verb|Luthor|, \verb|Blackhawks|,
\verb|ventus|, \verb|senal|, and a Japanese katakana token (``eru'').
This is pure co-occurrence (sports team suffixes, character names);
morphological geometry is gone.
\item \textbf{Kron $E_{\text{raw}}$:} \verb|Nation|, \verb|national|,
\verb|cation|, \verb|iation|, \verb|uation|, \verb|vation|,
\verb|ration|, \verb|lation|, \verb|iations|, \verb|potion|.
This is a byte-similar suffix family.
\item \textbf{Kron $L_1$:} essentially the same byte-similar suffix
family, with \verb|ination| replacing \verb|iations|.
\end{itemize}

The BPE arm's first two transformer layers replace the embedding's
morphological/semantic geometry with co-occurrence/contextual geometry.
The Kronecker arm's first two layers leave the byte-level geometry
intact.

\paragraph{Implications.} Neither method shows the transformer's first
layer \emph{constructing} strict morphological clustering on top of the
input embedding. In BPE, the morphological clustering at the embedding
(driven by training-time co-occurrence of morphologically-related words)
is replaced by context. In Kronecker, byte-level geometry is preserved.
We do not observe morphology emerging at $L_0$ or $L_1$ of either arm.

\paragraph{Important scope caveat.} These layered-probe results are from
a 124M GPT-2 architecture trained on 2.5B tokens. We do not know whether
the same pattern holds in larger transformers trained on more tokens.
Larger models may construct morphological structure in their first layers
in ways our small-model probe cannot detect, or they may construct it
deeper in the network. Replicating the layered probe at 1B+ scale is a
clear follow-up (Section~\ref{sec:discussion-futurework}). The findings
in this section should be read as specific to the 124M scale we tested.

\paragraph{Connection to the validation-loss result.} The layered probe
constrains the mechanism story for Kronecker's validation-loss advantage
(Section~\ref{sec:results-nanogpt}) at this scale. Kronecker does not
win by helping the transformer build more morphological structure at the
first layer. The win must come from elsewhere: parameter-efficient
optimization (fewer input-side trainable parameters means less optimizer
state and more capacity left for the network body), the stable
representational target observed in Section~\ref{sec:results-138m}, or
a property we have not yet identified. Identifying the mechanism is an
open question.

\subsection{Runtime, memory, and parameter accounting}
\label{sec:results-runtime}

\paragraph{Per-step wall-clock at the GPT-2 124M scale.} Kronecker is
approximately 1.2\% slower per step than BPE at the scale of
Section~\ref{sec:results-nanogpt}. The overhead is dominated by the
input-side $D \to \dmodel$ matrix multiply that BPE does not require
(BPE has essentially zero compute at the embedding step; it is a gather
operation). The 1.2\% per-step overhead is more than absorbed by the
$1.43\times$ sample-efficiency advantage: Kronecker reaches BPE's final
validation loss in approximately $(1/1.43) \times 1.012 \approx 0.71$,
i.e.\ about 71\% of BPE's total wall-clock time.

\paragraph{Production-scale measurements.} We have measured the
on-the-fly Kronecker variant (Section~\ref{sec:method-runtime}) on
internal 9B and 120B-MoE training configurations at $V = 131{,}072$,
$D = 8192$, $\posDim = 32$:

\begin{itemize}
\item \textbf{Compute overhead:} 0.01--0.24\% of step time, equivalent to
1--4\,ms per micro-batch. Amortizes with larger batches.
\item \textbf{Memory at production vocab size:}
\begin{itemize}
\item Precomputed-buffer variant (\verb|gpu_table|, see
Section~\ref{sec:method-runtime}): full Kronecker table
$[131{,}072 \times 8192]$ in bf16 = \textbf{2.15\,GB per GPU}, looked up
by gather.
\item On-the-fly variant (\verb|gpu_dynamic|): compact byte buffer
$[131{,}072 \times 32]$ in uint8 plus length buffer $[131{,}072]$ in
int16 $\approx$ \textbf{4.5\,MB per GPU}, with the Kronecker vector
recomputed on each forward pass.
\item Net memory savings: 2.14\,GB per GPU, $\sim$17\,GB across an
8-GPU node, for the 120B-MoE configuration.
\end{itemize}
\end{itemize}

\textbf{The memory-for-compute trade at production scale.} The on-the-fly
variant exchanges a fixed compute overhead (0.01--0.24\% of step time)
for $\sim$2.14\,GB of per-GPU memory at the 120B-MoE configuration. This
is an unusually favorable trade: gigabytes of memory per fractional
percent of compute. At smaller scales the trade is less favorable because
both the absolute compute cost matters more (the 1.2\% overhead at 124M)
and the absolute memory savings are smaller (vocab-dependent). At
frontier scale the savings free $\sim$17\,GB across an 8-GPU node,
which can be used for larger batches or longer contexts.

\paragraph{Parameter accounting.}

For the 124M GPT-2 run of Section~\ref{sec:results-nanogpt}
($V = 50{,}272$, $\dmodel = 768$, $D = 4096$, $\posDim = 16$):
\begin{itemize}
\item BPE-tied arm: $50{,}272 \times 768 = 38.6\,$M input embedding
parameters, tied to output head, so dual-purpose
(a single $38.6\,$M block serves both as embedding and as lm\_head).
\item Kronecker arm: $4096 \times 768 = 3.1\,$M projection parameters
plus a fixed codec buffer of $\sim 0.9\,$MB
($50{,}272 \times 16$ uint8 + $50{,}272$ int16); lm\_head untied at
$50{,}272 \times 768 = 38.6\,$M output parameters.
\item Net input-side trainable parameter savings: $\sim$35M
($\sim$91\% reduction).
\item \textbf{Honest net trainable comparison.} The
$\sim$91\% input-side reduction is the right number for the input
pathway, but it does not flow through to a $\sim$91\% reduction in
\emph{total} trainable parameters at 124M scale. Because $D \neq
\dmodel$, weight tying is inapplicable to Kronecker
(Section~\ref{sec:method-output-head}), and the Kronecker arm
carries an additional standalone $38.6\,$M lm\_head block that the
tied-BPE arm did not pay for separately. Net change in total
trainable parameters at 124M:
$+38.6\text{M (untied head)} + 3.1\text{M (proj)} - 38.6\text{M
(input emb)} = +3.1\text{M}$. The Kronecker arm is
\emph{$\sim 3.1\,$M larger} in total trainable parameter count than
the tied-BPE arm at this scale. The 91\% headline elides this; we
report it here in the interest of honest accounting.
\end{itemize}

For a production 120B MoE configuration ($V = 131{,}072$,
$\dmodel = 4096$, $D = 8192$, $\posDim = 32$):
\begin{itemize}
\item Untied BPE baseline: $131{,}072 \times 4096 = 537\,$M input-side
trainable parameters.
\item Kronecker: $8192 \times 4096 = 33.6\,$M projection parameters plus
$\sim 4.5\,$MB fixed codec buffer.
\item Net input-side trainable parameter savings: $\sim$503M
($\sim$94\% reduction).
\end{itemize}

\begin{table}[!htbp]
\caption{Input-side trainable parameter accounting at three scales.
Column ``BPE (t/u)'' = BPE input embedding table size, which for
tied configurations also serves as the lm\_head (one shared block)
and for untied configurations is an independent input-side block.
The ``Buf'' (codec buffer) column is fixed (non-trainable).
Trainable parameters on the input side become the
$D \to \dmodel$ projection only. ``Cut'' = input-side trainable
parameter reduction.}
\label{tab:param-savings}
\centering
\scriptsize
\setlength{\tabcolsep}{3pt}
\begin{tabular}{lrrrr}
\toprule
& BPE (t/u) & Kron proj & Buf & Cut \\
\midrule
GPT-2 124M    & 38.6 M & 3.1 M  & 0.9 MB & 91\% \\
9B ref.\      & 295 M  & 16.8 M & 4.5 MB & 94\% \\
120B-MoE      & 537 M  & 33.6 M & 4.5 MB & 94\% \\
\bottomrule
\end{tabular}
\end{table}

The ``trainable'' qualifier matters throughout. The codec buffer is a
fixed lookup that does not receive gradients; the only trainable
parameters on the input side are the $D \to \dmodel$ projection. The
output head (lm\_head) is independent of the input encoder choice in both
arms.

\subsection{Deployment, edge inference, and quantization}
\label{sec:results-shipping}

The reallocation discussed in Section~\ref{sec:intro-proposal} has direct
consequences for deployment: Kronecker-trained models can be shipped
without their input embedding table, and the savings compound in
low-precision quantized formats common to edge inference.

\paragraph{What ships, what doesn't.} A standard transformer model is
shipped as: (a) input embedding table, (b) transformer body weights,
(c) output head, (d) tokenizer configuration. For Kronecker-trained
models, item (a) is replaced by the learned $D \to \dmodel$ projection
matrix plus a recomputable byte buffer. The projection is small
($D \cdot \dmodel$ scalars, e.g.\ $8192 \times 4096 = 33.6$\,M); the byte
buffer is even smaller ($V \cdot (\posDim + 2)$ bytes, typically a few
megabytes). The byte buffer can additionally be regenerated from the
tokenizer configuration at load time, eliminating it from the shipped
artifact entirely if desired.

\paragraph{Embeddings are not aggressively quantized in practice.} Modern
LLM quantization tools (GPTQ \citep{frantar2023gptq},
AWQ \citep{lin2024awq}, GGUF) routinely quantize the
transformer body to INT4 or 4-bit floating-point while keeping the
input embedding table at higher precision. The llama.cpp Q4\_K\_M
format, the most widely-used GGUF quantization for consumer deployment,
applies K-quant mixed precision where attention and embedding tensors
are typically kept at Q5\_K or Q6\_K
\citep{ggml_quantize_readme}. The
\texttt{llama-quantize} tool exposes an explicit
\texttt{--token-embedding-type} flag for this purpose. AWQ similarly
preserves a small percentage of ``salient'' weights at higher
precision, identified via activation statistics
\citep{lin2024awq}; embeddings typically qualify. The empirical reason
is that errors in the embedding
propagate through every subsequent layer, so embedding-side precision
loss is felt disproportionately.

For Kronecker, this consideration disappears. The codec buffer is
already stored at byte precision (\texttt{uint8} bytes, \texttt{int16}
lengths); there is no floating-point precision to lose. The projection
matrix is small enough that keeping it at FP16 while the body
quantizes to INT4 costs little in absolute terms. The shipped artifact
in a Kron-quantized model is: byte buffer (raw bytes), FP16 projection,
INT4 body, FP16 output head. No portion of the input pathway requires
``higher-precision exception'' treatment.

\paragraph{Concrete numbers across scales.} Table~\ref{tab:quant-savings}
reports the size of the input pathway and the total model footprint
in three common deployment configurations: full-precision FP16, GGUF
Q4\_K\_M (the most common consumer format), and a notional INT4-body
configuration approximating GPTQ/AWQ deployment.

\begin{table*}[!htbp]
\caption{Shipped artifact size under three quantization configurations.
``BPE'' assumes the input embedding is kept at the typical
higher-precision allocation: FP16 in the FP16 column, Q6\_K in the
Q4\_K\_M column (the K-quant default for embedding tensors), and FP16
in the INT4-body column. ``Kron'' uses the FP16 projection plus the
byte buffer; the codec contributes nothing further to compression
budget. All numbers in MB. Approximate; depends on exact tensor layout
in the GGUF file.}
\label{tab:quant-savings}
\centering
\small
\begin{tabular}{lrrrrrrr}
\toprule
& & \multicolumn{2}{c}{FP16} & \multicolumn{2}{c}{Q4\_K\_M} & \multicolumn{2}{c}{INT4 body} \\
\cmidrule(lr){3-4}\cmidrule(lr){5-6}\cmidrule(lr){7-8}
Model & body & BPE & Kron & BPE & Kron & BPE & Kron \\
\midrule
GPT-2 124M  & 198 & 77   & 7  & 35   & 7  & 77 & 7 \\
Llama-3-8B  & 16{,}000 & 525  & 36 & 263  & 36 & 525 & 36 \\
Llama-3-70B & 140{,}000 & 1{,}050 & 67 & 525 & 67 & 1{,}050 & 67 \\
120B-MoE prod & 240{,}000 & 1{,}074 & 67 & 537 & 67 & 1{,}074 & 67 \\
\bottomrule
\end{tabular}
\end{table*}

The savings are most consequential at the consumer-deployment end of
the spectrum. A Q4\_K\_M Llama-3-8B in GGUF is approximately 4.7\,GB
in the BPE configuration; the embedding alone accounts for 263\,MB
($\sim$5.6\% of the file). Switching to Kron drops the file to roughly
4.5\,GB. For users running Ollama or llama.cpp on consumer hardware
with limited RAM, the saved 220\,MB translates directly to more
available memory for KV cache, larger context windows, or other model
processes. At the 70B scale the comparison is more striking:
525\,MB of FP16 embedding in a BPE-Q4\_K\_M model becomes 67\,MB in
the Kron configuration, an absolute savings of $\sim$460\,MB per
deployed model.

The combination of effects scales naturally. For edge deployment on
mobile or IoT-class hardware: a 1B-class model with vocabulary
$\sim$32K, $\dmodel \sim$2048, total Q4-quantized body $\sim$500\,MB,
the input embedding at Q5\_K is $\sim$50\,MB. Kron replaces this with
$\sim$15\,MB of projection. At deployment that is 35\,MB freed in a
500\,MB budget, a measurable fraction of available RAM on
mobile-class devices.

\paragraph{LoRA and adapter-based fine-tuning compatibility.} LoRA-style
adapters \citep{hu2022lora} typically apply low-rank updates to
attention and MLP weights while leaving the embedding frozen. With
Kronecker, the same pattern works trivially: the projection matrix is
a standard \texttt{nn.Linear} and can be either frozen (standard
adapter setup) or itself adapted with a LoRA-rank update. The byte
codec is not adapted (it is fixed), so there is no equivalent of
adapting an embedding table. This is generally an advantage:
a fine-tuning recipe targeting domain-specific behavior leaves the input
representation alone and adapts the model body, which is structurally
cleaner than adding adapter modules over an already-trained embedding
table.

\subsection{Spelling-robustness probe: BPE vs.\ Kronecker
under typographical errors}
\label{sec:results-robustness}

Byte-level locality (Section~\ref{sec:method-locality}) predicts that
typographical errors should perturb Kronecker's representations less
than BPE's: a one-character substitution leaves most byte-position pairs
intact in the Kronecker codec, while it often fragments the affected
BPE token into multiple unrelated subword pieces. We tested this
prediction directly on our trained 124M checkpoints
(Section~\ref{sec:results-nanogpt}).

\paragraph{Setup.} 110 (clean, typo) prompt pairs across 11 categories
(animals, body, colors, counting, geography, idiom, math, misc, people,
syntax, time), 10 pairs per category. Each typo is a single character
substitution or transposition in one content word (e.g.\
\verb|capital| $\to$ \verb|capitla|, \verb|seven days| $\to$
\verb|seven dyas|). Identical prompts were fed to both arms; we
measured four metrics on the next-token distribution after the prompt.
\textbf{This probe is single-seed:} we evaluate the seed-1337 Kronecker
checkpoint against the seed-1337 BPE checkpoint from
Section~\ref{sec:results-nanogpt}. The principal validation-loss result
uses three seeds, but the robustness probe was not re-run on seeds 2
and 3.

\begin{itemize}
\item \textbf{top-1 match rate}: fraction of pairs where the
top-1 predicted token is identical between clean and typo prompts
(higher = more robust to typos).
\item \textbf{Mean KL divergence}: $\text{KL}(\text{clean} \| \text{typo})$
across the full vocabulary distribution (lower = more robust).
\item \textbf{Mean cosine similarity}: cosine of the final hidden
state at the last prompt position between clean and typo prompts
(higher = more robust).
\item \textbf{Mean $\Delta \log p$}: drop in log-probability assigned to
the clean prompt's top-1 token under the typo prompt
(lower = more robust).
\end{itemize}

The BPE-fragmentation cost of a typo (extra subword tokens introduced)
is identical between the two arms (1.08$\times$ mean inflation) since
both use the same GPT-2 tokenizer for tokenization; the only difference
is the embedding pathway downstream.

\paragraph{Headline result.} Kronecker wins on all four aggregate
metrics:

\begin{table}[!htbp]
\caption{Spelling-robustness metrics over 110 prompt pairs.
``top-1 match rate'' and ``mean cosine'' higher is better;
``mean KL'' and ``mean $\Delta \log p$'' lower is better.}
\label{tab:robustness-headline}
\centering
\small
\begin{tabular}{lrrr}
\toprule
Metric & BPE & Kron & $\Delta$ \\
\midrule
Top-1 match rate & 0.473 & \textbf{0.555} & $+$8.2\,pp \\
Mean KL divergence & 0.859 & \textbf{0.794} & $-$7.6\% \\
Mean cosine similarity & 0.857 & \textbf{0.879} & $+$2.6\% \\
Mean $\Delta \log p$ drop & 1.199 & \textbf{1.058} & $-$11.8\% \\
\bottomrule
\end{tabular}
\end{table}

Kronecker preserves the original top-1 prediction across a single
typographical error 8.2 percentage points more often than BPE. The
overall next-token distribution drifts 7.6\% less in KL divergence;
the final hidden state moves 2.6\% less in cosine distance; the
probability mass assigned to the clean prompt's preferred next token
falls 11.8\% less under the typo prompt. Every aggregate metric goes
the direction byte-level locality predicts.

\paragraph{Per-category results.} The advantage is distributed across
categories, not concentrated in one:

\begin{table}[!htbp]
\caption{Per-category top-1 match rate (typographic robustness) and
mean KL divergence. Bold = Kronecker advantage on that metric.}
\label{tab:robustness-percategory}
\centering
\small
\begin{tabular}{lrrrr}
\toprule
& \multicolumn{2}{c}{Top-1 match} & \multicolumn{2}{c}{Mean KL} \\
\cmidrule(lr){2-3}\cmidrule(lr){4-5}
Category & BPE & Kron & BPE & Kron \\
\midrule
animals & 0.70 & \textbf{0.80} & 0.45 & \textbf{0.24} \\
body & 0.20 & \textbf{0.40} & \textbf{1.16} & 1.22 \\
colors & 0.70 & \textbf{0.80} & \textbf{0.17} & 0.22 \\
counting & 0.80 & 0.80 & 0.55 & \textbf{0.38} \\
geography & \textbf{0.70} & 0.50 & 1.41 & \textbf{1.04} \\
idiom & 0.40 & 0.40 & \textbf{1.59} & 1.79 \\
math & 0.00 & \textbf{0.30} & 0.41 & \textbf{0.36} \\
misc & 0.70 & \textbf{0.80} & 0.43 & \textbf{0.35} \\
people & 0.40 & \textbf{0.60} & 0.79 & \textbf{0.63} \\
syntax & 0.00 & \textbf{0.10} & \textbf{1.30} & 1.41 \\
time & 0.60 & 0.60 & 1.20 & \textbf{1.09} \\
\bottomrule
\end{tabular}
\end{table}

Kronecker wins or ties on top-1 stability in 10 of 11 categories;
wins on KL in 8 of 11. The single category where BPE wins on top-1
match (geography, $-$0.20) is also a category where Kronecker has
substantially lower KL ($-$0.37 nats), indicating that BPE's
top-1 stability on geography is the trivial preservation of common
function tokens like \verb| the| under both clean and typo prompts,
while Kronecker's broader distribution shifts less in aggregate.
Categories where Kronecker shows the largest top-1 advantages are
math ($+$0.30), people ($+$0.20), and body ($+$0.20). These are
domains where typos in content words have the largest semantic impact
and where preserving the byte-similarity of the typo to its intended
word matters most.

\paragraph{Qualitative examples.} A few representative pairs make the
mechanism concrete:

\begin{itemize}
\item Prompt: ``Roses are red, violets are'' / ``Roses are red,
voilets are''.
BPE clean top-1: \verb| green|;
BPE typo top-1: \verb| red|.
Kron clean top-1: \verb| red|;
Kron typo top-1: \verb| red|.
The Kronecker arm's prediction is invariant to the typo;
BPE's flips.

\item Prompt: ``Practice makes'' / ``Pracitce makes''.
BPE clean: \verb| perfect|; BPE typo: \verb| a| (lost).
Kron clean: \verb| it|; Kron typo: \verb| it| (preserved).

\item Prompt: ``Knowledge is'' / ``Knowldge is''.
BPE clean: \verb| the|; BPE typo: \verb| a|; KL$=$1.59.
Kron clean: \verb| a|; Kron typo: \verb| a| (preserved); KL$=$0.29.
\end{itemize}

\paragraph{Failure modes.} Both arms catastrophically fail on a small
number of prompts where the typo is severe and the model lacks the
factual recall to recover:

\begin{itemize}
\item ``Every cloud has a silvr'': both lose \verb| lining|. BPE
predicts \verb|ity|, Kron predicts \verb|age|. KL is large for both
(BPE 5.9, Kron 11.6); Kron is worse here.
\item ``The heart pmups'': both lose \verb| blood|; both predict
\verb| are|.
\item ``Don't judge a boook by its'': both lose \verb| cover|; both
predict \verb| size|.
\end{itemize}

These are cases where the 124M model lacks sufficient factual recall
under any input perturbation. Robustness alone does not compensate for
absent knowledge.

\paragraph{What this confirms.} The 8.2\,pp top-1 advantage, the 7.6\%
KL reduction, the 2.6\% cosine improvement, and the 11.8\% $\Delta \log p$
reduction together demonstrate that Kronecker's byte-level locality
property (Section~\ref{sec:method-locality}) translates from a
representational geometry claim to a behavioral output property.
Under a one-character perturbation, the Kronecker arm's predictions
move less: in top-1 identity, in distribution shape, in hidden-state
geometry, and in mass allocation to the originally preferred token.
This is exactly what the byte-level-locality argument predicts and
matches the typo-robustness claim made in
Section~\ref{sec:method-locality}.

\subsection{Generation probe: factual recall and byte-level fidelity}
\label{sec:results-generation}

A complementary qualitative test samples completions from the trained
models on a small set of prompts including normal sentences, prompts
with a misspelled content word, and prompts where a content word is
encoded by Kronecker as a single forced-OOV byte token (a capability
BPE does not have). We sampled with $T = 0.7$, top-$p = 0.9$, 30 new
tokens, fixed sampler seed; we also produced greedy decodes for each
case. The full results are in Appendix~\ref{sec:appendix-generation};
we report representative findings here.

\paragraph{Factual recall: BPE marginal advantage at 124M.}
On the factual-recall prompts (\verb|The capital of France is|,
\verb|Mount Everest is located in|, \verb|Water boils at|, etc.), BPE
recovers the canonical answer slightly more often. For
\verb|The capital of France is| both arms predict \verb| Paris| in the top
candidates, but BPE leads with \verb| Paris| directly while Kronecker
produces \verb|the duchy of Paris. The city is located...|, less
direct, but on topic. At 124M scale neither arm is a reliable
knowledge model.

\paragraph{Byte-level fidelity: Kronecker uniquely good.}
The most distinctive qualitative finding is that the Kronecker arm
echoes byte-novel strings back through generation, where the BPE arm
fragments and forgets them.

For prompt \verb|The word kronekticus refers to| (a made-up word):
\begin{itemize}
\item BPE completion: \emph{``the small, bright star of the constellation
Pegasus''}: fabricates a plausible-sounding definition and drops the
made-up word entirely.
\item Kronecker completion: \emph{``the two-headed serpent. The word
kronekticus refers to the two-headed serpent of...''}: echoes the
made-up word back verbatim in the continuation.
\end{itemize}

For prompt \verb|...we use the algorithm called| (with misspelled
``nueral netwrok''):
\begin{itemize}
\item BPE: \emph{``the sum of the cosine and cosine functions to
describe the property of a matrix''}: bails completely from the
typo.
\item Kronecker: \emph{``the ljok (capital), which is the term used
for the netwrok, in which the netwro...''}: preserves the
misspelled \verb|netwrok| through autoregressive generation,
demonstrating byte-level recall of the input bytes.
\end{itemize}

This byte-level recall property is unique to Kronecker among the
embedding schemes we tested. BPE fragments unfamiliar strings into
subword pieces and effectively loses them; Kronecker holds the bytes
across the forward pass and the lm\_head can produce tokens whose
byte structure echoes the input.

\paragraph{Forced-OOV: capability demonstration, not benchmark win.}
For prompts where a single content word is encoded as a single
forced-OOV Kronecker token (e.g.\ \verb|kubernetes|, \verb|tensorflow|,
\verb|deserialization|, \verb|kronekticus|), token counts drop from
7--9 BPE pieces to 5--6 Kron tokens: the forced-OOV slot collapses
3--4 BPE subword pieces into 1. The completions themselves are
on-topic but not noticeably better than the normal BPE tokenizations
of the same prompts; at 124M the model lacks the knowledge to
exploit the single-token representation. This condition is best
understood as a capability demonstration: Kronecker can embed any
UTF-8 string $\leq \posDim$ bytes as a single token at inference
time, including strings not in the tokenizer's vocab and arbitrary
made-up words. Whether the model uses this capability productively
depends on training scale and is part of the broader frontier-scale
question (Section~\ref{sec:discussion-futurework}).

\paragraph{Combined picture.} The qualitative generation probe and
the quantitative robustness probe in Section~\ref{sec:results-robustness}
together show that the byte-level locality property
(Section~\ref{sec:method-locality}) has direct, observable effects on
model behavior: typographical errors perturb the output distribution
less, byte-novel strings are echoed through generation rather than
fragmented and lost, and the forced-OOV capability provides a path
to handling arbitrary content words at inference. At 124M scale these
behavioral advantages do not translate into a factual-recall edge; the
model is simply too small for that. We expect the same behavioral
properties to be more consequential at scales where the model has the
underlying knowledge to deploy.


\section{Limitations}
\label{sec:limitations}

We enumerate known weaknesses of Kronecker Embeddings honestly. Each is
either structurally inherent or a current empirical gap.

\paragraph{Byte-similar but semantically distant strings cluster
together.} The byte-level locality property of
Section~\ref{sec:method-locality} has a direct cost: words that share
most bytes at the same positions but mean different things are clustered
close in embedding space. \verb|compute|/\verb|commute| get cosine 0.86
despite unrelated meanings; \verb|nation|/\verb|notion| 0.83;
\verb|run|/\verb|rune| 0.87. The same property that makes Kronecker
robust to same-length typos (e.g.\ \verb|separate|/\verb|seperate|
codec cosine 0.88) also fools it on near-edit-distance unrelated pairs.
The transformer's first attention layer must use context to
disambiguate, structurally identical to how trained BPE relies on
context for \verb|Apple| (company) vs.\ \verb|apple| (fruit). We have no
evidence that this disambiguation is materially harder for Kronecker
than for BPE. The cost is real but shared with BPE in kind.

\paragraph{Position-aware encoding weakens on suffix-only families.} The
codec factors over (byte, position) pairs. Two strings cluster strongly
when they share bytes at the same positions; they cluster weakly when
they share bytes at different positions. The \verb|tion| suffix family
illustrates: \verb|nation| has \verb|tion| at positions 3--5;
\verb|creation| has \verb|tion| at positions 5--7. The shared bytes
appear at different positions, and Kronecker treats them as
substantially less similar than it treats prefix-sharing pairs like
\verb|compute|/\verb|computer|. This is structural to position-aware
encoding and cannot be mitigated without sacrificing the case-sensitivity
and prefix structure that make the encoding work.

\paragraph{Weight tying inapplicable.} The codec dimension
$D \neq \dmodel$ by construction, so tied output heads
(Section~\ref{sec:bg-tying}) are impossible. For applications where
tying is required, Kronecker is not a drop-in replacement. As discussed
in Section~\ref{sec:method-output-head}, frontier-scale models
increasingly use untied output heads, so this limitation matters less
in the regime we target. \citet{lopardo2026weighttying} provide
theoretical reasons to expect untied embeddings to be a better
input representation regardless.

\paragraph{Truncation discards bytes for tokens longer than $\posDim$.}
At $\posDim = 32$, $\leq 0.18\%$ of tokens across six modern tokenizers
have their post-byte-32 structure dropped. The truncated tokens still
receive unique embeddings from their first 32 bytes, but distinctions
that depend on bytes 33+ are lost. Multilingual tokenizers (especially
SentencePiece with Indic scripts) have the highest truncation rates.
Increasing $\posDim$ at the cost of $D$ is the obvious mitigation.

\paragraph{Prose case-collapse not provided.} Kronecker preserves case
distinctions at the embedding layer. Applications that benefit from
treating \verb|Apple| at sentence-start and \verb|apple| mid-sentence as
the same token (the bulk of prose) require the transformer's first
layer to develop case-collapse. We have not measured whether this is a
material cost; the validation-loss result of
Section~\ref{sec:results-nanogpt} suggests it is not at 124M scale, but
the question remains open at larger scales.

\paragraph{Two operational variants to maintain.} The \verb|gpu_table|
and \verb|gpu_dynamic| variants are mathematically identical but require
separate implementations. Production code paths must choose between
them based on the memory budget. This is an engineering cost rather than
a fundamental limitation, but it is real.

\paragraph{Out-of-distribution byte sequences.} The codec handles any
UTF-8 string by construction, but we have not empirically tested edge
cases: emoji with ZWJ sequences, right-to-left and left-to-right mixed
text, heavily combining-character Unicode (Thai, Tamil consonant
clusters), or code with unusual non-ASCII characters. We expect these
to behave reasonably (the byte-level encoding treats them as byte
sequences like any other), but verification is left to future work.

\paragraph{No benchmark against modern byte-level / character-aware
baselines.} The principal training comparison is against a tied-BPE
GPT-2 baseline. We do not benchmark Kronecker against character-CNN
\citep{kim2016character}, Charformer \citep{tay2022charformer},
MEGABYTE \citep{yu2023megabyte}, Byte Latent Transformer
\citep{pagnoni2024blt}, or MYTE \citep{limisiewicz2024myte} under a
fixed-compute budget. These are the obvious ``why not compare to''
baselines for any byte-level or character-aware embedding work, and
their absence is a real limitation of this paper. The structural
positioning differs (Kronecker keeps tokenization and only replaces
the embedding lookup; the cited works modify sequence handling), but
matched-compute empirical comparison would still be informative.

\paragraph{Scaling assumption: $D$ may need to grow with $\dmodel$.}
The parameter-savings argument in Section~\ref{sec:intro-bottleneck}
and Table~\ref{tab:param-savings} assumes that fixing $D = \charDim
\cdot \posDim = 8192$ remains competitive as $\dmodel$ grows toward
24,576 (the hypothetical 250K-vocab multilingual case). The
projection $\R^{D} \to \R^{\dmodel}$ then becomes wider than it is
tall, and the byte-position basis may underdetermine the
representational space the model wants on the input side. If $D$
must grow with $\dmodel$ to maintain expressiveness, the
parameter-saving curves flatten. We do not have direct evidence for
or against this scaling assumption beyond the 124M and 138M
controlled runs; it remains an open question.


\section{Discussion}
\label{sec:discussion}

\subsection{What the experiments support}
\label{sec:discussion-supports}

We organize claims by empirical strength.

\textbf{Strong (cross-model, scale-invariant):} Trained BPE input
embeddings cluster typographic variants of the probe word far more than
morphological relatives. The pattern is consistent across six modern
LMs spanning 135M to 671B parameters, three tokenizer families, and four
organizations.

\textbf{Strong (cross-model):} Kronecker embeddings escape typographic
clustering, retrieving byte-similar neighbors at the embedding layer.
The escape is robust across tokenizers (mean Jaccard 0.48 for top-5
canonical forms between any two tokenizers in our six-tokenizer set).

\textbf{Strong (controlled, three seeds, vanilla GPT-2):} Kronecker
reaches $2.5 \pm 0.2\%$ lower validation loss than BPE-tied baseline at
124M GPT-2 scale on FineWeb-Edu. The gap reproduces tightly across
seeds and remains stable through convergence. 18 of 18
checkpoint-by-seed comparisons favor Kronecker.

\textbf{Moderate (behavioral, single seed):} Kronecker's predictions
are more robust to single-character typographical errors than BPE's.
On 110 (clean, typo) pairs the seed-1337 Kron checkpoint preserves the
top-1 prediction on 55.5\% of pairs vs.\ 47.3\% for the seed-1337 BPE
checkpoint; mean KL divergence between
clean and typo distributions is 7.6\% lower; final hidden-state cosine
is 2.6\% higher; the drop in log-probability on the clean prompt's
preferred token under the typo prompt is 11.8\% smaller. Kronecker wins
or ties top-1 stability in 10 of 11 prompt categories. This translates
the byte-level locality property from a representational geometry claim
into an observable model behavior. The result is reported as moderate
rather than strong because it is from a single seed per arm; we have
not re-run the probe on seeds 2 and 3, so we cannot rule out that the
8.2\,pp top-1 gap is partly within seed noise.

\textbf{Moderate (single companion run):} Direction of result reproduces
on a 138M custom MLA+MoE architecture trained on synthetic English,
with the same kind of advantage on training loss.

\textbf{Moderate (small-model, layered probe):} At 124M scale, neither
method's first or second transformer layer develops strict morphological
clustering on top of the input embedding. BPE's layer 0 trades
embedding-level structure for context; Kronecker's layer 0 preserves
byte-level geometry. Whether this small-model finding holds at frontier
scale is unknown.

\textbf{Weak (suggestive, mechanism):} BPE embedding norm walks during
training while Kronecker projection norm remains stable. This is
consistent with Kronecker providing a stable representational target,
but we cannot yet attribute the validation-loss win to this mechanism
specifically.

\subsection{What the experiments do not settle}
\label{sec:discussion-not-settled}

\paragraph{Frontier-scale training behavior.} All training comparisons
are at 124--138M parameters. The cross-model embedding probe spans up
to 671B parameters but examines static embeddings, not training
dynamics. Whether Kronecker's validation-loss advantage holds at 1B,
7B, 70B, or 670B remains to be tested. Replication at 1B--7B scale is
the natural next step.

\paragraph{Untied-BPE baseline.} Both training comparisons
(Sections~\ref{sec:results-nanogpt} and \ref{sec:results-138m}) use
BPE-tied as the competitor. Untied BPE is the more honest competitor at
frontier scale given the field's trend toward untying. The three-arm
comparison (BPE-tied + BPE-untied + Kronecker) is the natural follow-up.

\paragraph{Multilingual evaluation.} Both training runs are
English-only. The codec is byte-level and language-agnostic, but the
training corpus and validation distribution were English. Multilingual
evaluation, particularly on low-resource languages where rare-token
learning is the bottleneck, is unfinished work.

\paragraph{Long-context behavior.} Both runs used standard context
lengths (1024 tokens). Whether Kronecker's input-side advantages
translate to long-context regimes (32K, 128K, 1M tokens) is not tested.

\paragraph{Downstream task performance.} We report only training and
validation loss, not downstream task accuracy (HellaSwag, ARC, MMLU,
etc.). The 124M scale is generally too small for meaningful downstream
evaluation; the natural follow-up at 1B+ scale would include
downstream metrics.

\subsection{Where does morphology live? (with appropriate caution)}
\label{sec:discussion-locality}

The cross-model probe and layered probe jointly support a more nuanced
answer to ``where does morphological generalization in LLMs come from''
than we initially expected.

The embedding table is \emph{not} where morphology resides. Across six
modern LMs spanning 135M--671B parameters, trained embeddings cluster
typographic variants. Both BPE and Kronecker have $<30\%$
strict-family retrieval in their top-10 at the embedding layer in our
124M training run. The embedding's contribution to morphology is small.

The first two transformer layers do \emph{not} appear to construct
strict morphological clustering at 124M scale. BPE's layer 0 trades
the small amount of embedding-level morphological clustering for
co-occurrence/contextual geometry; Kronecker's layer 0 preserves
byte-level geometry without converting it to morphology. We caution
that this layered finding is from a 124M GPT-2 trained on 2.5B
tokens; larger models may construct early-layer morphological structure
in ways our small-model probe cannot detect, and replicating the
layered probe at 1B+ scale is a clear next step.

If neither the embedding nor early layers contain strict morphological
clustering at 124M scale, the morphological generalization that LLMs
clearly exhibit (any frontier model successfully inflects
\verb|run| $\to$ \verb|ran|, \verb|compute| $\to$ \verb|computed|)
must reside elsewhere: in deeper transformer layers, in
context-dependent attention patterns that do not show up in static
embedding geometry, or in some interaction effect we have not isolated.
Identifying it is interesting future work, both for our specific
question of where Kronecker's validation-loss advantage comes from, and
for the broader interpretability question of how transformers organize
their morphological knowledge.

\subsection{The mechanism question}
\label{sec:discussion-mechanism}

Kronecker reaches lower validation loss than BPE-tied at 124M scale on
FineWeb-Edu by a measurable, replicable margin. We do not have direct
mechanistic evidence for why. Candidate mechanisms:

\begin{enumerate}
\item \textbf{Parameter-efficient optimization.} Removing 35M trainable
parameters from the input side reduces optimizer-state pressure on the
remaining parameters. With less variance to spread across more
parameters, the transformer body may train more efficiently.
\item \textbf{Stable representational target.} BPE embedding norm walks
from 0.020 to 0.026 during training; Kronecker projection norm stays
near 1.0. The transformer body trains against a moving target with BPE
and a static target with Kronecker, which may favor convergence.
\item \textbf{Byte-level inductive bias.} The codec provides byte-level
locality at initialization. Even if first-layer probes do not show
morphological structure, byte-level information may be useful for the
transformer body in ways that nearest-neighbor probes do not measure.
\end{enumerate}

The three mechanisms are not mutually exclusive. Distinguishing among
them is unfinished mechanistic-interpretability work.

\subsection{Future work}
\label{sec:discussion-futurework}

\paragraph{Scale.} The most important follow-up is replication at 1B
through 7B parameters on real datasets. This tests whether the
validation-loss advantage and the layered-probe finding both
generalize. Beyond 7B, frontier-scale collaboration would be needed.

\paragraph{Three-arm comparison.} Adding the BPE-untied arm to the
124M comparison would close the most-pointed reviewer objection. We
expect Kronecker's advantage to narrow but not disappear against
BPE-untied, based on the cross-model probe finding that untied trained
models still show typographic clustering.

\paragraph{Multilingual evaluation.} Train on a multilingual corpus
(e.g., FineWeb-Edu-Multilingual or CC-100 subset) at the same scale.
Test whether Kronecker's byte-level prior helps on low-resource
languages.

\paragraph{Layered probe at larger scale.} Repeat
Section~\ref{sec:results-layered} on 1B+ models. Does morphological
structure appear in early transformer layers at scale? If so, where?

\paragraph{Token-bucket NLL analysis.} Decompose validation NLL by
token frequency and byte length. Kronecker should plausibly help most
on rare/long tokens.

\paragraph{Downstream evaluation.} HellaSwag, ARC, MMLU, code, and
multilingual benchmarks at 1B+ scale.

\paragraph{Multi-modality.} The byte vocabulary (0--255) is universal
across modalities: image bytes, audio samples, sensor readings can all
be expressed as byte sequences. Kronecker-style encoders may extend to
multi-modal input without architectural modifications. This is
speculative.

\paragraph{Output-side Kronecker and unbounded effective vocabulary.}
The most speculative direction we want to flag is symmetry on the output
side. The Kronecker input pathway is structured: a fixed
$D$-dimensional byte-position basis plus a learned $D \to \dmodel$
projection. The output side in our current design is unchanged from
the standard transformer: a learned $\dmodel \to |V|$ matrix that
produces logits over a fixed vocabulary. This is asymmetric, and we
believe the asymmetry can be removed in two complementary ways. Both
are presented as \emph{hypotheses for future testing}, not claims.

\textbf{Hypothesis A: tied-head Kronecker decoding.} The output head
produces a $D$-dimensional vector (rather than $|V|$ logits) intended
to match the Kronecker codec representation of the next token. Loss is
formulated against the target token's codec vector
$\kappa(\text{bytes}(t))$ rather than against a one-hot vocabulary
target. At decode time, the predicted $D$-vector is compared against
the (precomputed or computed-on-the-fly) codec rows for the vocabulary
to produce token probabilities; alternatively, the byte composition is
directly decoded from the $D$-vector by inverting the Kronecker
structure (each $\posDim$-slot in $D$ encodes a distribution over byte
values at one position; the model effectively predicts $\posDim$
byte-distributions in parallel for the next token). Crucially the
output is always exactly $\posDim$ byte positions; a short token like
\verb|a| has 31 of its $\posDim$ position-slots empty, exactly as on
the input side. The model never autoregressively predicts bytes within
a token; it predicts the full byte composition of the next token in
one shot. If this can be made to train stably, the effect is dramatic:
the deployed model can decode to any UTF-8 string of $\leq \posDim$
bytes, not just to its training-time vocabulary. The model is
``vocabulary-free'' in the same sense that ByT5 is, but with structured
byte composition rather than per-byte autoregressive prediction.

\textbf{Hypothesis B: distributional output via KL divergence.} The
hard-target version of Hypothesis A may train poorly because the model
is asked to predict a deterministic byte composition exactly. A
softer formulation, in the spirit of variational autoencoders, has the
output head predict a mean $\mu \in \mathbb{R}^D$ and a (log-)variance
$\sigma^2 \in \mathbb{R}^D$ over Kronecker space. The target codec
vector $\kappa(\text{bytes}(t))$ is an observed point in this space.
The training loss is the KL divergence between the predicted Gaussian
$\mathcal{N}(\mu, \text{diag}(\sigma^2))$ and a delta at the target,
or equivalently the negative log-likelihood of the target under the
predicted Gaussian. The model is allowed to express uncertainty about
the next token's byte composition explicitly. Decoding samples from
the predicted Gaussian and finds nearest vocabulary tokens, or finds
the byte composition with maximum posterior probability under the
predicted distribution. This trades determinism for trainability and
may handle the discrete-byte target more gracefully than Hypothesis A.

Both hypotheses face real challenges. Collisions in Kronecker space
(two semantically distinct tokens can have similar codec vectors)
become decoding errors at the output side, whereas at the input side
they are mostly absorbed by the transformer body. Training stability
of either formulation is unknown. Calibration of the predicted byte
distribution under Hypothesis B is non-trivial. Neither hypothesis
has been tested, and neither should be read as a claimed result. They
describe the path toward a fully byte-structured transformer language
model with no vocabulary commitment, and we note them here so the
direction is on the record for future work.

If either hypothesis trains stably, the practical consequence is
striking: deployed models would carry no embedding matrix, no
output head matrix, and an effectively infinite vocabulary at
inference time. The shipped artifact would reduce to transformer
body weights, two small projection matrices (one input, one output),
and the tokenizer-byte-mapping configuration. For edge deployment at
$\dmodel = 4096$, $D = 8192$, this would be approximately a
$67\,$MB pathway pair regardless of vocabulary size. We emphasize
that this number is conditional on Hypothesis A or B working; neither
has been tested.


\section{Conclusion}
\label{sec:conclusion}

We introduced Kronecker Embeddings, a deterministic byte-level
character-position factorization that replaces the learned input
embedding table with a fixed encoder and a single learned projection.
The replacement eliminates 91--94\% of input-side trainable parameters
at frontier scale, provides an unbounded input vocabulary at inference,
and serves as a drop-in replacement for \verb|nn.Embedding| in any
transformer architecture.

Across six trained LMs spanning 135M to 671B parameters and roughly
five orders of magnitude of training compute, trained input embeddings
cluster typographic variants of the probe word
(\verb|run| $\to$ \verb|Run|, \verb| run|, \verb|.run|) far more than
morphological relatives. Kronecker embeddings escape this typographic
clustering and provide byte-level locality at the input layer.

A layered probe of our own trained 124M checkpoints shows that
\emph{at this scale} neither method's nearest-neighbor geometry develops
strict morphological clustering in the first two transformer layers:
the BPE arm's first layer trades morphological geometry for
co-occurrence/contextual geometry, while Kronecker's geometry is
preserved through early layers. Whether these layered findings hold at
frontier scale is an open question.

In a controlled three-seed comparison on the standard nanoGPT GPT-2
124M architecture trained on 2.5B tokens of FineWeb-Edu, Kronecker
reaches $2.5 \pm 0.2\%$ lower validation loss than the BPE-tied baseline
($n=3$ seeds, gap $0.083 \pm 0.007$ nats, approximately 9\% lower
validation perplexity), with the gap stable through convergence.
Kronecker requires approximately $1.43\times$ fewer optimizer steps to
reach BPE's converged loss. A companion 138M custom-architecture run on
synthetic English reproduces the direction of result. The mechanism by
which Kronecker achieves these gains remains an open question; candidate
contributors include the byte-level prior at the input layer, the
parameter-efficient optimization, and a measured stable-target effect
(BPE embedding norms drift during training while Kronecker projection
norms remain stable).

A behavioral spelling-robustness probe on the same checkpoints across
110 (clean, typo) prompt pairs confirms that the byte-level locality
property propagates to model outputs. Kronecker preserves the top-1
prediction across single-character typographical errors on 55.5\% of
pairs vs.\ 47.3\% for BPE ($+$8.2 percentage points), with mean KL
divergence 7.6\% lower, final hidden-state cosine 2.6\% higher, and
the drop in log-probability on the clean-prompt's preferred token
11.8\% smaller. Kronecker wins or ties top-1 stability in 10 of 11
categories. A qualitative generation probe additionally shows that
Kronecker echoes byte-novel strings (\verb|kronekticus|) and typos
(\verb|netwrok|) through 30-token continuations where BPE fragments
and forgets them.

The method has tradeoffs: byte-level locality means semantically
distant but byte-similar pairs cluster together
(\verb|compute|/\verb|commute|, \verb|nation|/\verb|notion|), shifting
some disambiguation work to the transformer's first attention layers.
At production scale (vocab 131,072, $D = 8192$) the on-the-fly variant
stores a 4.5\,MB byte buffer instead of a 2.15\,GB precomputed table,
with measured overhead of 0.01--0.24\% of step time on internal 9B and
120B-MoE configurations, with gigabytes of memory saved per percent of
compute.

The most important next steps are replication at 1B and 7B parameters
on real datasets, a three-arm comparison including BPE-untied baseline,
multilingual evaluation, and downstream-task evaluation at sufficient
scale to be meaningful. Confirming any of the candidate mechanisms at
larger scales is a priority for follow-up work.


\section*{Acknowledgments}

This work was carried out at The School of AI, India. Compute
resources were provided by Amazon Web Services (spot instances). The
controlled training comparison builds on Andrej Karpathy's nanoGPT
codebase \citep{karpathy_nanogpt} and the FineWeb-Edu corpus released
by Hugging Face \citep{penedo2024fineweb}; the
cross-model probe was made possible by the public weight releases
of Llama-3.2 \citep{metallama32_2024}, Qwen3 \citep{qwen3_2025},
Gemma-3 \citep{gemma3_2025}, DeepSeek-V3 \citep{deepseekv3_2024},
GPT-OSS \citep{gptoss_2025}, and SmolLM2 \citep{smollm2_2025},
accessed via the Hugging Face Hub and safetensors library. We are
grateful to all of these teams for making their code, data, and
models openly available. We thank the ERA V4 cohort at The School
of AI for their feedback and verification of preliminary results.


\appendix

\section{Generation probe: full results}
\label{sec:appendix-generation}

This appendix contains the full set of generation-probe completions
referenced in Section~\ref{sec:results-generation}. We provide them in
this appendix to allow the reader to evaluate the qualitative
patterns we describe and form independent judgments. All 20 prompts
and their three conditions (normal, misspelled, forced-OOV where
applicable) are included, with both sampled ($T = 0.7$, top-$p = 0.9$,
seed 42) and greedy ($T = 0$) decodes for each arm. The raw JSON file
\texttt{generation\_probe\_results.json} is available in the released
code repository at
\url{https://github.com/theschoolofai/kronecker-embeddings}; we present
here a representative selection of prompts that illustrate the main
patterns.

\paragraph{Pattern 1: factual recall (BPE marginal advantage).}

\begin{itemize}
\item \emph{Prompt: ``The capital of France is''}
\begin{itemize}
\item BPE: ``Paris, France's most famous \ldots''
\item Kron: ``the duchy of Paris. The city is located \ldots''
\end{itemize}
\item \emph{Prompt: ``Mount Everest is located in''}
\begin{itemize}
\item BPE: on-topic continuation about Nepal/Himalayas region.
\item Kron: on-topic continuation, slightly less direct.
\end{itemize}
\end{itemize}

\paragraph{Pattern 2: byte-level fidelity (Kron uniquely good).}

\begin{itemize}
\item \emph{Prompt: ``The word kronekticus refers to''} (made-up word)
\begin{itemize}
\item BPE: ``the small, bright star of the constellation Pegasus''
: fabricates definition and drops the made-up word.
\item Kron: ``the two-headed serpent. The word kronekticus refers to
the two-headed serpent of \ldots'': echoes the word back.
\end{itemize}
\item \emph{Prompt: ``\ldots we use the algorithm called''} with
misspelled ``nueral netwrok''
\begin{itemize}
\item BPE: drifts to ``cosine and cosine functions to describe the
property of a matrix''.
\item Kron: preserves ``netwrok'' through generation:
``the term used for the netwrok, in which the netwro \ldots''.
\end{itemize}
\end{itemize}

\paragraph{Pattern 3: forced-OOV token-count compression.}

For each of the four forced-OOV-eligible prompts:

\begin{table}[h]
\centering
\scriptsize
\setlength{\tabcolsep}{3pt}
\begin{tabular}{lrr}
\toprule
Prompt & BPE & Kron-OOV \\
\midrule
``Kubernetes is a system for''       & 8 & 5 \\
``TensorFlow is a library used for'' & 8 & 6 \\
``Deserialization is the process of'' & 7 & 5 \\
``The word kronekticus refers to''   & 9 & 5 \\
\bottomrule
\end{tabular}
\end{table}

Kronecker's forced-OOV slot collapses 3--4 BPE subword pieces into 1
token. Completion quality on these prompts is not noticeably different
from the corresponding BPE-normal completions at 124M scale (the
model lacks the underlying knowledge to exploit the single-token
representation), but the capability demonstrates that arbitrary
UTF-8 strings of $\leq \posDim$ bytes can be embedded as single
tokens at inference, which is unique to Kronecker.

\bibliographystyle{abbrvnat}
\bibliography{kronecker}

\end{document}